\documentclass[preprint,12pt,authoryear]{elsarticle}

\usepackage{amssymb}

\usepackage{float}
\usepackage{tikz}
\usepackage{subcaption}
\usepackage{caption}
\usepackage{amsmath}
\usepackage{booktabs}
\usepackage{multirow}

\begin{document}

\begin{frontmatter}



\title{Comparing remote sensing-based forest biomass mapping approaches using new forest inventory plots in contrasting forests in northeastern and southwestern China}


\author[inst1]{Wenquan Dong\corref{cor1}}
\cortext[cor1]{Corresponding author}
\affiliation[inst1]{organization={School of GeoSciences, University of Edinburgh},
            addressline={King's Buildings}, 
            city={Edinburgh },
            postcode={EH9 3FF}, 
            country={UK}}
            
\author[inst1,inst2]{Edward T.A. Mitchard}
\affiliation[inst2]{organization={Space Intelligence Ltd},
            addressline={93 George Street}, 
            city={Edinburgh },
            postcode={EH2 3ES}, 
            country={UK}}         

\author[inst3]{Yuwei Chen}
\affiliation[inst3]{organization={Department of Remote Sensing and Photogrammetry, Finnish Geospatial Research Institute},
            addressline={Geodeetinrinne 2}, 
            city={Kirkkonummi},
            postcode={02430}, 
            country={Finland}}  
            
\author[inst1]{Man Chen\corref{cor1}}

\author[inst5]{Congfeng Cao}
\affiliation[inst5]{organization={Institute for Logic, Language and Computation, University of Amsterdam},
            addressline={Science Park 900}, 
            city={Amsterdam},
            postcode={1098 XH, NL}, 
            country={Netherlands}}
            
\author[inst3,inst6]{Peilun Hu}
\affiliation[inst6]{organization={Department of Forest Science, University of Helsinki},
            addressline={Yliopistonkatu 4}, 
            city={Helsinki},
            postcode={00100}, 
            country={Finland}}

\author[inst7]{Cong Xu}
\affiliation[inst7]{organization={Jihua Lab},
            addressline={No 28 huandao south road}, 
            city={Foshan},
            postcode={528200}, 
            country={China}}            

\author[inst1]{Steven Hancock}            
            
\begin{abstract}
Large-scale high spatial resolution aboveground biomass (AGB) maps play a crucial role in determining forest carbon stocks and how they are changing, which is instrumental in understanding the global carbon cycle, and implementing policy to mitigate climate change. The advent of the new space-borne Light Detection and Ranging (LiDAR) sensor, NASA's Global Ecosystem Dynamics Investigation (GEDI) instrument, provides unparalleled possibilities for the accurate and unbiased estimation of forest AGB at high resolution, particularly in dense and tall forests, where Synthetic Aperture Radar (SAR) and passive optical data exhibit saturation. However, GEDI is a sampling instrument, collecting disperate footprints, and its data must be combined with that from other continuous cover satellites to create high-resolution maps, using local machine learning methods. In this study, we developed local models to estimate forest AGB from GEDI L2A data, as the models used to create GEDI L4 AGB data incorporated minimal field data from China. We then applied an ensemble model of decision trees (Light Gradient Boosting Machine, LightGBM), and random forest regression to generate wall-to-wall AGB maps at 25 m resolution, using extensive GEDI footprints as well as Sentinel-1 C-band data, Advanced Land Observing Satellite (ALOS) -2 Phased Array L-band SAR (PALSAR) -2 data and Sentinel-2 optical data. We found many GEDI footprints had errors not detected by their quality flags; filtering using remote sensing data to find realistic possible values improved results. Through a 5-fold cross-validation, LightGBM demonstrated a slightly better performance than Random Forest across two contrasting regions. In the northeastern region, the coefficient of determination (R$^2$) of the two models is both 0.70 and the root-mean-square-error (RMSE) ranges between 28.44 and 28.77 Mg ha\textsuperscript{-1}. Meanwhile, in the southwestern region, the R$^2$ values range from 0.60 to 0.62, and the RMSE values range between 37.71 to 38.29 Mg ha\textsuperscript{-1}. However, in both regions, the computation speed of LightGBM is substantially faster than that of the random forest model, requiring roughly one-third of the time to compute on the same hardware. Through the validation against field data, the 25 m resolution AGB maps generated using the local models developed in this study exhibited higher accuracy compared to the GEDI L4B AGB data. We found in both regions an increase in error as slope increased. In the northeastern region, the R$^2$ values decreased from 0.93 in flat areas to 0.42 in areas with \textgreater30 degrees slope, while the RMSE increased from 14 to 38 Mg ha\textsuperscript{-1}. In the southwestern region, the R$^2$ values decreased from 0.75 in flat areas to 0.50 in areas with \textgreater30 degrees slope, while the RMSE increased from 20 to 44 Mg ha\textsuperscript{-1}. The trained models were tested on nearby but different regions and exhibited good performance. This study not only proposes an approach to map forest AGB using extensive GEDI and continuous SAR, optical data in China but also enables the estimation of AGB changes using the trained model, even in the absence of GEDI but presence of SAR and optical data.

\end{abstract}

\begin{highlights}

\item Unique field data collected co-located with GEDI footprints to estimate forest aboveground biomass from GEDI L2A with locally tuned models.
\item New filters using SAR data proved to be highly beneficial for extracting high-quality GEDI footprints.
\item Compared gradient boosting ensemble machine learning algorithm and random forest 
\item The trained models performed well in other areas of China.

\end{highlights}


\end{frontmatter}


\section{Introduction}

Forests play an essential role in the health and sustainability of the planet's ecosystems \citep{watson2018exceptional}. Accurate information on forest aboveground biomass (AGB) is critical for effective forest management, carbon accounting, and climate change mitigation efforts \citep{xu2021changes, houghton2009importance}. Traditional forest inventory methods for estimating forest biomass are labor-intensive and time-consuming, making them impractical for large-scale monitoring and assessment. Approaches based on earth observations have emerged as a promising alternative for estimating forest AGB \citep{mitchard2018tropical, tucker2023sub}, offering the potential to rapidly and consistently map AGB over large areas. Optical, SAR, and LiDAR are the three main types of earth observation data.

Passive optical data measures light reflected from the Earth's surface, and is widely used due to the ability to capture detailed spectral information, which can be used to assess canopy cover and potentially estimates of AGB. Optical data provides the longest temporal record,  with the series of Landsat that commenced data collection in 1972. It has traditionally been an efficient tool to map wall-to-wall forest AGB by combining with field data through regression models \citep{mcnicol2018carbon, lu2004relationships, zhu2015improving, h2019landsat, zheng2004estimating}. However, optical data can only see the top of the forest canopy, meaning similar canopy with different heights or trunk widths, are not be distinguished well. Additionally, optical data saturates at low AGB values and its utility can be significantly hampered by atmospheric conditions such as clouds and haze. 

SAR is a type of active sensor that transmits its own energy and subsequently records the quantity of that energy reflected back \citep{meyer2019spaceborne}. The long wavelengths of the SAR sensors used for biomass mapping (C, L, and longer wavelengths if available) enables it to see through clouds, which ensures consistent data acquisition. The continuous, reliable, all-weather, day-and-night imaging capabilities of SAR make it particularly useful for forest AGB estimation \citep{mitchard2009using}. SAR can further penetrate through the canopy and interact with leaves, branches and trunks, and thus SAR is sensitive to forest AGB \citep{balzter2007forest}. SAR data which has longer wavelength tends to have a higher saturation point than those with shorter wavelengths \citep{woodhouse2012radar, sinha2015review}. However, even the L-band SAR, which has the longest wavelength among currently available space-borne SAR data, has a relatively low saturation point for AGB at around 100-150 Mg ha\textsuperscript{-1} \citep{mitchard2009using, englhart2011aboveground, mitchard2012mapping}.

LiDAR data employs laser light to estimate tree height and structure, thereby facilitating the construction of a detailed representation of a forest. This technology allows for the direct estimation of 3D structural parameters such as height, stem density, and canopy cover, providing invaluable insights into forest characteristics and dynamics. Therefore, LiDAR suffers little from saturation effects \citep{wulder2012lidar, duncanson2022aboveground}, although it cannot penetrate through clouds or haze. With the launch of GEDI, the high-resolution LiDAR instrument which is designed to measure vegetation structure \citep{dubayah2020global}, it is possible to quantify forest AGB in distributed 25m diameter footprints, a footprint size chosen to minimize slope effects compared to the previous spaceborne LiDAR the Ice Cloud and Land Elevation Satellite (ICESat) with its 65m footprints, over very large areas.

However, useful instruments do not necessarily produce accurate forest AGB maps. China encompasses diverse forest cover types and complex terrain, making it a challenging and highly variable environment for AGB estimation \citep{dai2011major}. China is among the countries with the highest degree of inaccuracy when it comes to estimating AGB using GEDI data. The AGB estimated by GEDI for China is more than double the estimates provided by the Food and Agriculture Organization (FAO) \citep{dubayah2022gedi} and other studies \citep{su2016spatial, huang2019integration, chang2021new}. One potential reason for this is the lack of training data in China \citep{duncanson2022aboveground}. Forest biomass estimation using remote sensing relies on the correlation between remotely sensed data and ground-based biomass measurements, which vary greatly depending on the forest type and terrain conditions such as elevation and slope. In this study, we collected new field plots co-located with GEDI footprints to develop relationships to convert GEDI relative height (RH) metrics into forest AGB.

However, unlike SAR and optical data, GEDI data are not continuous. The discontinuous LiDAR data is often combined with optical and SAR data to generate wall-to-wall estimates \citep{saatchi2011benchmark}. Their combined use can provide a more comprehensive and accurate estimation of forest carbon. Random forest \citep{breiman2001random} is one of the most popular algorithm to extrapolate LiDAR footprints to wall-to-wall map using optical and SAR imagery \citep{dong2024new, baccini2012estimated, huang2019integration, narine2020using}.

In this study, we used random forest as well as a more advanced machine learning algorithm, LightGBM \citep{ke2017lightgbm}, to generate wall-to-wall AGB maps with a resolution of 25 m. LightGBM is a gradient boosting ensemble method that is known for its efficiency and ability to handle large-scale data \citep{yan2021lightgbm}. However, the universality of the two models across different different regions is not yet fully understood. We tested the performance of the two models in two contrasting regions, and validated the results using the field data. In addition, the trained models were applied to respective areas. The remote sensing data used in the models were from GEDI, Sentinel-1, Sentinel-2, and ALOS-2, all of which are open-source. We aim to propose a universally applicable method for estimating AGB in China, or if necessary, demonstrate the need for region-specific models across different regions.
  
\section{Data and methods}
\subsection{Study area}
\begin{figure}[H]
    \centering
    \begin{subfigure}{1.0\textwidth}
        \centering
        \begin{tikzpicture}
            \node[anchor=south west,inner sep=0] (image) at (0,0) {\includegraphics[width=\textwidth]{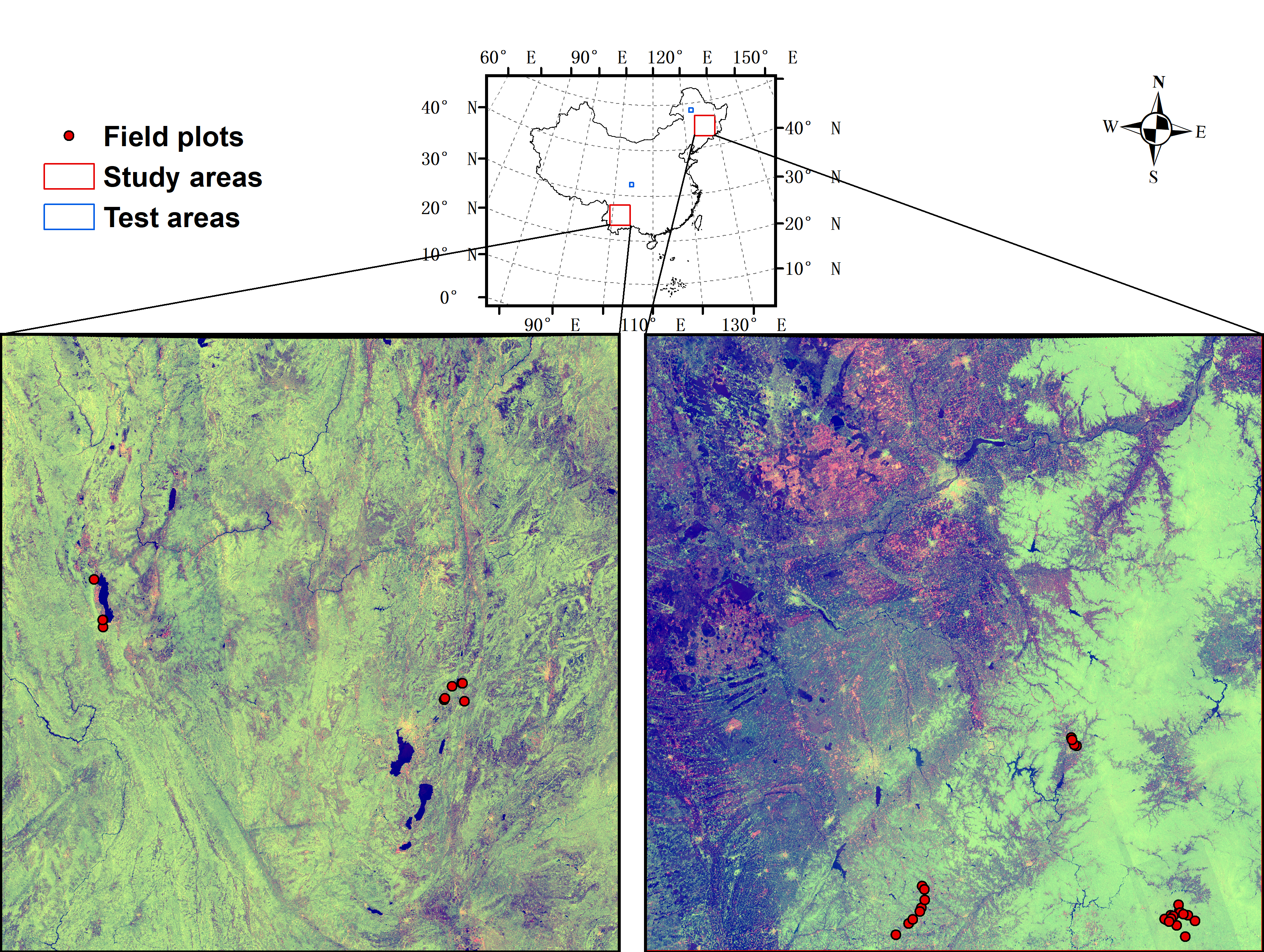}};
        \end{tikzpicture}
    \end{subfigure}%

\caption{Location of the study area and field sites. The field plots are shown on the composited ALOS PALSAR-2 mosaic (HH in red, HV in green, and HV/HH in blue).}
\label{fig: Study area}
\end{figure}

This study was conducted in two contrasting areas of China: one in northeastern China, and the other one in southwestern China (Fig. \ref{fig: Study area}). Each of the study areas covers an area of 500 x 500 km, but the types of forests and terrain are completely different. The forest types differ in terms of their structure, species composition, and environmental conditions, providing an ideal opportunity to test the universality of remote sensing-based forest biomass mapping approaches which is of great significance for achieving large-scale forest AGB predictions in areas with complex forest compositions.

\begin{figure}[H]
    \centering
    \begin{subfigure}{1.0\textwidth}
        \centering
        \begin{tikzpicture}
            \node[anchor=south west,inner sep=0] (image) at (0,0) {\includegraphics[width=\textwidth]{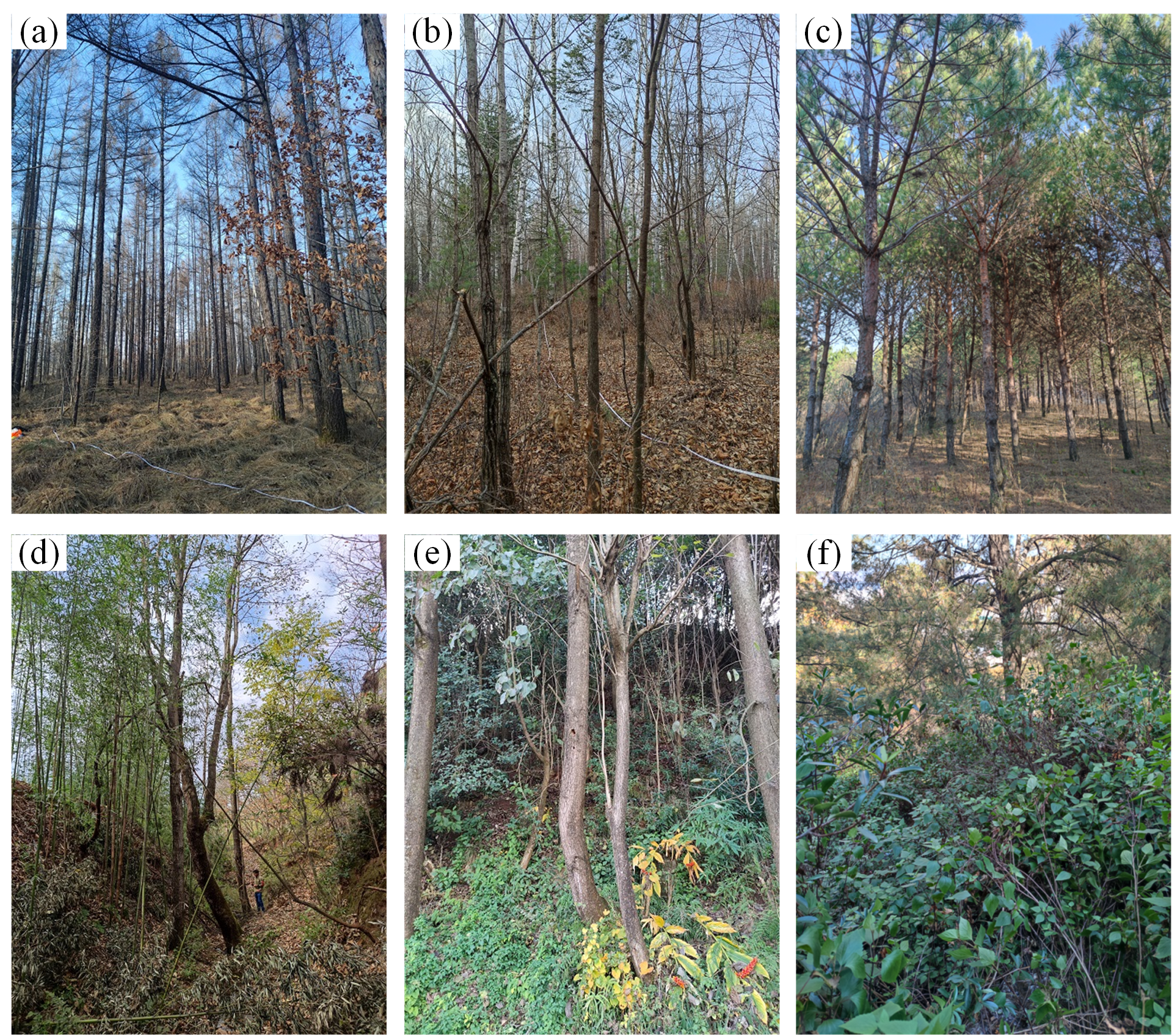}};
        \end{tikzpicture}
    \end{subfigure}%

\caption{Field photos showing different forest types and terrains in the two study areas.}
\label{fig: Field figure}
\end{figure}

The northeastern study area, primarily located in Jilin Province as well as the southern part of Heilongjiang Province, is characterized by the temperate forest. The area experiences a continental monsoon climate, with long, cold winters and short, warm summers \citep{liu2018spatial}. The average temperatures in January and July are -20 and 21 $^\circ$C, respectively. The average annual precipitation ranges from 600 to 1100 mm \citep{dong2005forest}. The forests are dominated by Korean pine (Pinus koraiensis) mixed with deciduous species (e.g. Betula platyphylla) \citep{wang2006biomass} (Fig. \ref{fig: Field figure}). A portion of the primary forest has been destroyed by large-scale industrial logging, and replaced by secondary forests and plantations \citep{yu2011forest}, which mainly consist of Betula platyphylla, Larix gmelinii, and Pinus koraiensis \citep{shi2015effects}.

The southwestern study area is mainly situated in Yunnan Province. Forests in this region predominantly blanket the mountainous terrain, as this area is dominated by plateau and mountainous areas, with elevations ranging from 285 to 5350 m, and an average elevation of 2039 m (Digital Elevation Model). The average temperature fluctuates between roughly 9–11 $^\circ$C in January and climbs to around 22 $^\circ$C in July, while the average annual precipitation varies from 1100 to 1600 mm \citep{shi2018characteristics}. The natural conditions described above contribute to the richness and diversity of Yunnan's forest resources, including needleleaf forest, subtropical evergreen broad-leaved forests and deciduous broadleaf forest \citep{zhang2019forest, zhu2022environmental, frayer2014processes}.

\subsection{Field data of AGB}
The field measurements were carried out during October and December 2021. We measured 26 plots in northeastern region with 24 plots were under GEDI footprints, and 16 plots in southwestern region with 12 plots were under GEDI footprints. For the collection of field data, we used the geolocation of the GEDI Version 1 footprints.The geolocation uncertainty associated with these footprints is reported to be 23.8 meters \citep{roy2021impact}, which may have introduced some degree of uncertainty in the co-localization of GEDI footprints and field plots. In order to align with the GEDI data, all field plots were circular with a diameter of 25 m. We endeavored to maintain a substantial distance between field points (typically greater than 2 km) and ensured they were at least 50 meters away from non-forest areas. This strategy was adopted to avoid issues related to mixed pixels.

After identifying the GEDI footprints to be measured (Fig. \ref{fig: Study area}), we navigated to the center of each footprint using a Garmin eTrex 20x handheld GPS device. To improve the accuracy of our measurements, we averaged the GPS readings over a period of 5 minutes. A tape was then used to delineate a circular plot, centered on the core of the footprint, with a diameter of 25 meters. Subsequently, we measured the height and Diameter at Breast Height (DBH) of trees within the plot that have a DBH $\geq$ 5 cm. This methodology ensures a comprehensive and accurate assessment of the field plot. DBH was measured at 1.3 m up the trunk from the ground using a diameter tape. For the majority of the field plots, tree height was estimated using a Vertex IV with a Transponder T3. However, for two plots in the northeastern region, the tree height was measured using a clinometer. The field AGB of individual tree was then obtained using tree height and DBH, calculated via the following  allometric equation \citep{luo2018china}.
\begin{equation}
W = 0.1355*(D^2*H)^{0.817}
\label{eq: allometric}
\end{equation}
where, $W$ is the AGB of each tree (kg), $D$ and $H$ are DBH (cm) and height (m) of the tree. We sum the AGB of all trees within the plot that have a DBH $\geq$ 5cm. This total AGB was then converted to a per hectare basis for standardization, expressed in Megagrams per hectare (Mg ha\textsuperscript{-1}).

\subsection{LiDAR data}

\subsubsection{Canopy height metrics from GEDI}
The GEDI instrument captures the waveform of the light that has bounced back from the Earth's surface to the GEDI sensor on the International Space Station, with a 25 m spatial footprint every 60 m along orbits. The GEDI instrument is equipped with three lasers. Out of these, two lasers operate at full power while the third one is divided into two coverage beams. These four beams are dithered across-track, creating a total of eight ground tracks of data, comprising four full-power beams and four coverage beams \citep{dubayah2020global}.

The dataset essentially presents the distribution of Earth's surface features, like vegetation and terrain, within the footprint of a laser shot. Specifically, it includes details on the vertical structure of the features, which is of great importance when assessing characteristics like canopy height or the vertical distribution of foliage in forests \citep{dubayah2020global}. In this study, L2A dataset acquired from 2019 to 2022 was used to produce continuous forest AGB map. 

\subsubsection{Airborne laser scanning data}
Airborne laser scanning (ALS) are capable of collecting data on vegetation structure, which can potentially serve as a proxy for traditional field measurements \citep{potapov2021mapping}. We employed ALS point clouds, collected in the summer of 2022, with a point density of 18 points$/m^2$  in our research. The ALS data consists two areas, the first area of 18 ha and the other area of 22 ha, co-located with 11 GEDI footprints respectively.

\subsection{SAR data}
To obtain as much information as possible on forest structure, we utilized both C-band and L-band SAR in this study. 

\subsubsection{Sentinel-1 C-band data}
Sentinel-1 Interferometric Wide swath (IW) product acquired during 2020 and 2022 was used in the estimation. Sentinel-1 was accessed and processed in Google Earth Engine (GEE). The Sentinel-1 Ground Range Detected (GRD) data had been calibrated and terrain-corrected, and the values were converted to decibels. 

The backscatter coefficient of VV (vertical-vertical) and VH (vertical-horizontal) polarizations was exploited, as different polarizations carries information about the structure of the forest \citep{flores2019sar}. Their cross ratio (VV/VH) was also included, as the cross ratio of Sentinel-1 is likely to mitigate errors associated with the acquisition system \citep{veloso2017understanding} and has been demonstrated to effectively differentiate between vegetation densities \citep{vreugdenhil2018sensitivity}. To reduce the impact of temporal variations and anomalies that might occur in individual images, we calculated temporal means of the data. To further reduce the speckle noise, we applied a focal mean speckle filter with a radius of 50 m to the data. 

\subsubsection{ALOS-2 PALSAR-2 L-band data}
The annual mosaics of L-band SAR data from ALOS-2 PALSAR-2 was utilized in this study. The 25 m resolution yearly mosaic dataset was created by the Japan Aerospace Exploration Agency (JAXA), using data acquired in the target year. The ortho-rectified and slope corrected data was accessed in GEE, and we also applied a focal mean speckle filter with a radius of 50 m to the means of annual mosaic of 2020 to 2022 to reduce the impact of speckle noise. The digital numbers (DN) of HV and HH were then converted to gamma naught values in decibel using the the following equation \citep{shimada2009palsar}.
\begin{equation}
\gamma_0 = 10\log_{10}(\text{DN}^2) - 83.0 \, \text{dB}
\label{eq:palsar}
\end{equation}
where, $\gamma_0$ is the backscatter coefficient in dB. In addition to the both polarizations, we also prepared the ratio (HV/HH) data, which has been identified as effective strategies for enhancing the saturation point \citep{hayashi2019aboveground,sarker2012potential}.

\subsection{Optical data}
Optical data were also used to extrapolate LiDAR footprints to wall-to-wall image. The MultiSpectral Instrument (MSI) on board the Sentinel-2 satellite measures reflected radiance across 13 spectral bands. These bands span a range from the Visible and Near Infrared (VNIR) to the Short Wave Infrared (SWIR) spectra. This broad coverage enables detailed observations of various surface conditions and phenomena. The ortho-rectified and atmospherically corrected L2A surface reflectance (SR) dataset, excluding band 10, is publicly accessible in GEE. In addition to utilizing the mean values of each band of L2A imagery acquired between 2020 and 2022, we also computed the mean and maximum values of the Normalized Difference Vegetation Index (NDVI), which provide valuable information regarding the average and peak level of vegetation growth.

\subsection{Topographic data}
The NASA Digital Elevation Model (NASADEM) is a global digital elevation model with a resolution of 30 m. NASADEM combines data from different satellite missions, including the Shuttle Radar Topography Mission (SRTM), the Advanced Spaceborne Thermal Emission and Reflection Radiometer (ASTER), and ICESat, to generate a seamless global elevation model \citep{jpl2020nasadem}. In this study, we utilized the terrain slope derived from NASADEM as an additional variable for analysis. The terrain slope was computed using a 3 × 3 moving window approach. The integration of terrain slope information provided valuable insights into the topographic characteristics of the study areas, enhancing  the impact of topographic variations on the AGB predictions.

\subsection{Forest AGB estimation}
\subsubsection{GEDI data filtering} \label{sec:GEDI data filtering}

Before proceeding with model fitting, we implemented multiple filters to identify and remove poor-quality footprints. L2A footprints collected from 2019 to 2022 were filtered using the following criteria \citep{dubayah2021gedi}.

(a) Only shots that have a quality\_flag of 1 were used. This flag signifies that the shot meets a summation of energy level, sensitivity, amplitude, quality of real-time surface tracking, and the difference from DEM.

(b) Shots acquired during degrade period were excluded.

(c) Only shots with a sensitivity $>$ 0.98 were used. A higher sensitivity indicates a greater likelihood of penetrating to the ground beneath a specific level of canopy cover.

(d) Shots were included if they were acquired at night, when the negative impact of solar illumination on the quality of GEDI waveform is eliminated, thus yielding better tree height prediction results \citep{liu2021performance}.

\begin{figure}[H]
    \centering
        \centering
        \begin{tikzpicture}
            \node[anchor=south west,inner sep=0] (image) at (0,0) {\includegraphics[width=\textwidth]{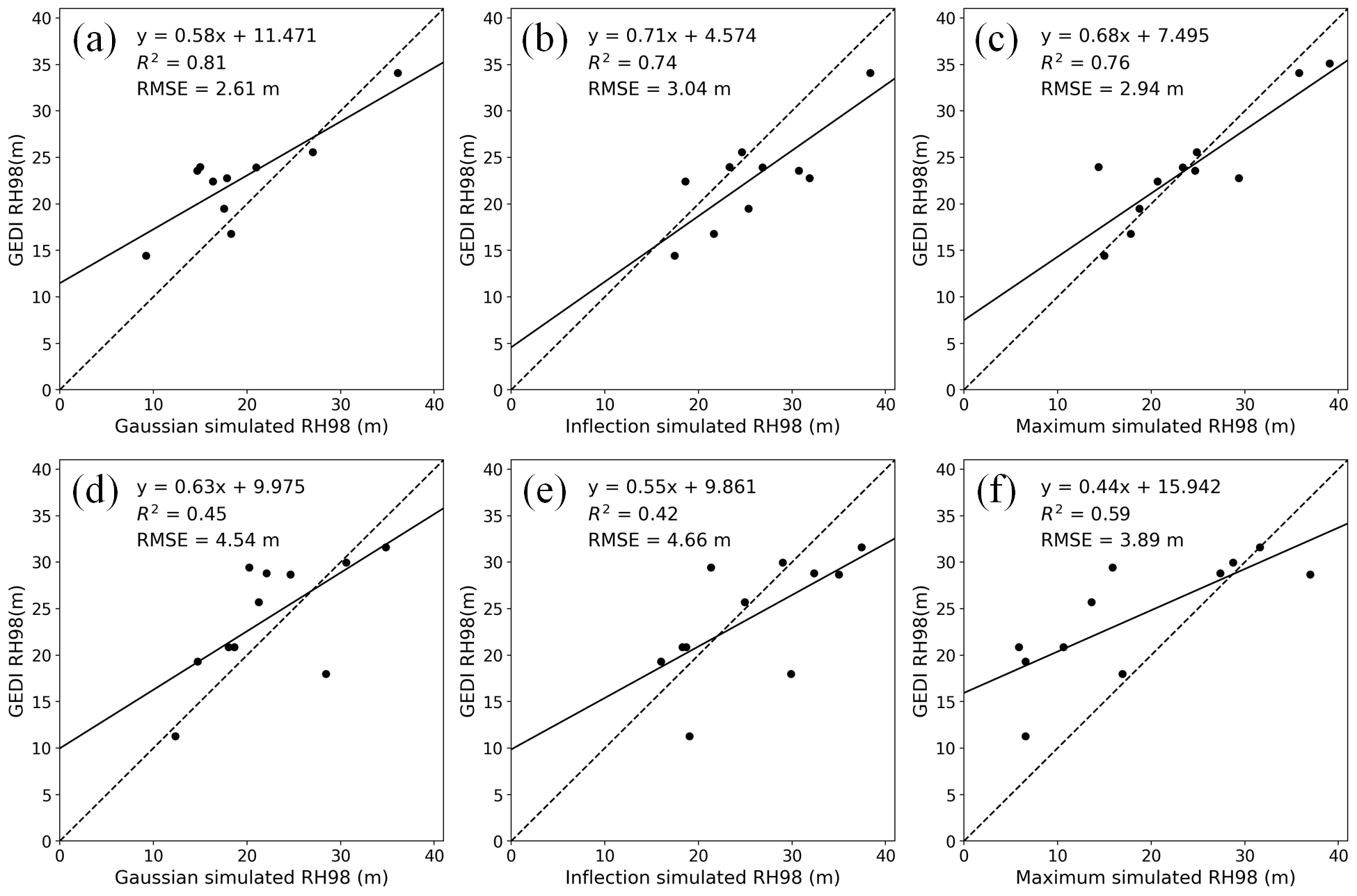}};
        \end{tikzpicture}
\caption{RH98 of actual GEDI against GEDI simulator. (a)-(c) is the scatter plot for the first area, all the footprints are comprised within full-power beam. (d)-(f) is the scatter plot for the second area, all the footprints are comprised within coverage power beam. The dotted line represents 1:1 line. The solid line represent the fitted line, and is extended to cover the entire range of the x-axis. This was done to provide a comprehensive visual representation of the relationships between GEDI RH metrics and ALS simulated RH metrics.}
\label{fig:GEDI against ALS}
\end{figure}

After we filtered the GEDI footprints, we then compared GEDI L1B data with ALS data to verify the quality of GEDI data, and to eliminate poor-quality footprints in GEDI. In order to compare with GEDI footprints, we used ALS data to correct geolocation errors in GEDI observations and to simulate GEDI data using three ground-finding algorithms: the Gaussian fitting, the lowest inflection point and the lowest maximum \citep{hancock2019gedi}.The simulated GEDI observations from ALS data was then compared with actual GEDI footprints (fig. \ref{fig:GEDI against ALS}).

As expected, footprints of GEDI power beams have a higher relationship with ALS simulated footprints, with a higher R$^2$ and a smaller RMSE. Since this also had been proved in other study areas \citep{duncanson2020biomass, li2023first,  liu2021performance}, we only included power beams in this study.

By applying the approach proposed by \citet{hancock2019gedi} to compare GEDI and ALS data, we found that the horizontal geolocation errors of GEDI data in China are substantial, with a 17.9 m error for the first study area and a 9.8 m error for the second area. We noticed a significant difference in the geolocation error between the two areas, indicating that the geolocation error of some footprints is larger than others. This means that the RH metrics at the actual geolocation of some GEDI footprints may differ significantly from the RH metrics at the geolocation it shows, which could introduce large error into AGB predictions. 

Because of this finding, we proceeded with an additional filtering process on the GEDI footprints using ALOS-2 PALSAR-2 HV, HH polarizations, and Sentinel-1 VH, VV polarizations, which have been proven to have a close relationship with forest height and AGB \citep{mitchard2012mapping, duncanson2020biomass, cartus2012mapping, ge2022improved, santi2017potential}. We fitted the relationship between GEDI RH98 and SAR imagery in the Northeast region. The reliability of the fitting relationships were assessed using R$^2$ and RMSE, calculated using the following equations:

\begin{equation}
R^2 = 1 - \frac{\sum_{i=1}^{n}(SAR_i - f(RH_i))^2}{\sum_{i=1}^{n}(SAR_i - \overline{SAR})^2}
\end{equation}
\begin{equation}
RMSE = \sqrt{\frac{1}{n}\sum_{i=1}^{n}(SAR_i - f(RH_i))^2}
\end{equation}
where $SAR_i$ is the value of SAR imagery, $f(RH_i)$ is the fitted value, $overline{SAR}$ is the  average value of $SAR_i$. The R$^2$ and RMSE here can be influenced by the fitted function.

\begin{figure}[H]
    \centering
        \centering
        \begin{tikzpicture}
            \node[anchor=south west,inner sep=0] (image) at (0,0) {\includegraphics[width=\textwidth]{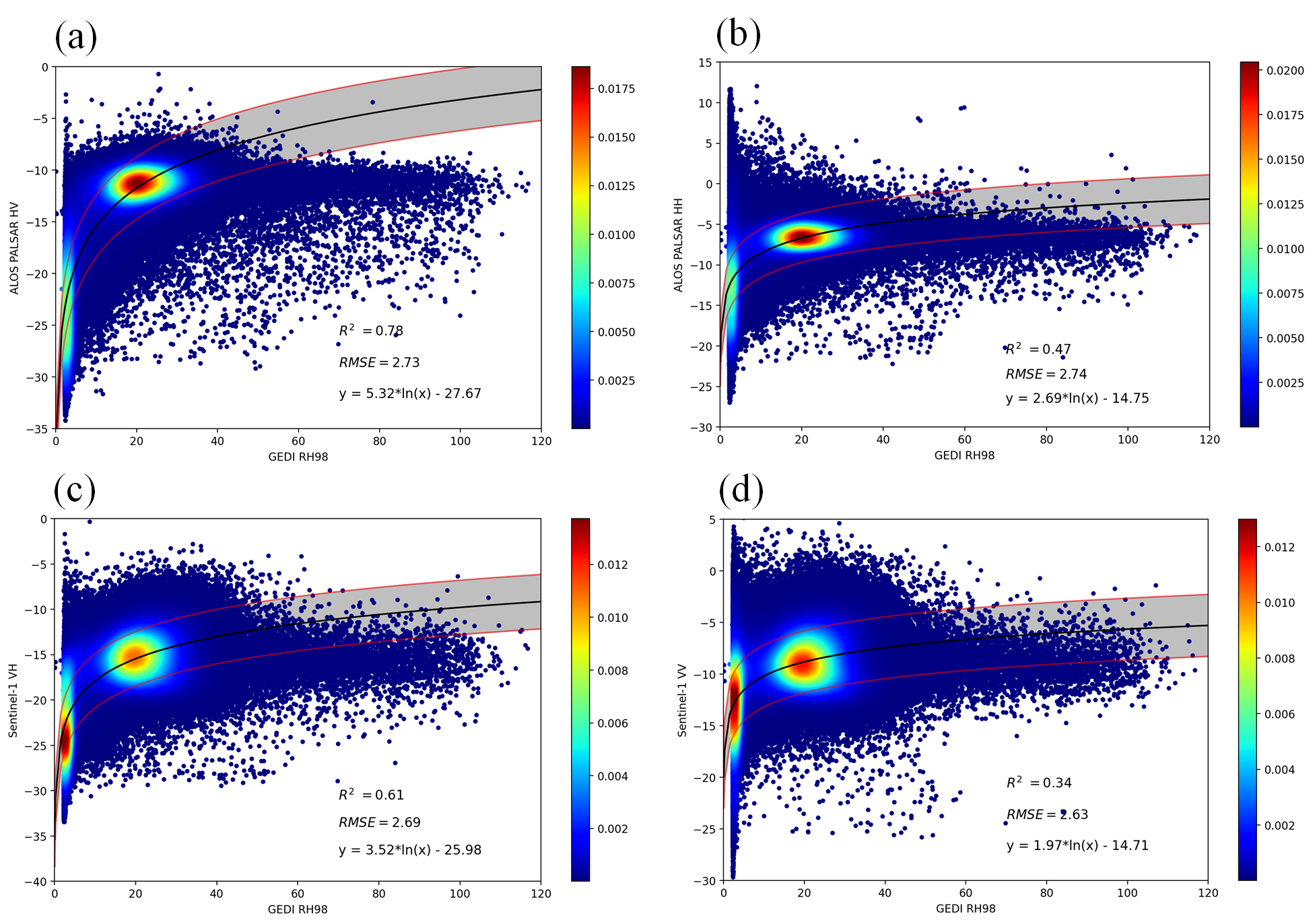}};
        \end{tikzpicture}
\caption{ALOS PALSAR-2 backscatter in (a) HV, (b) HH polarizations, and Sentinel-1 backscatter in (c) VH, (d) VV polarizations were plotted against GEDI RH98 for 1,405,451 GEDI footprints in the northeastern region. The black solid line represents the fitted line. The red solid line above the black solid line represents the fitted line plus 3 dB, and the red solid line below the black solid line represents the fitted line minus 3 dB. The color of each point in the scatter plot represents the estimated density of points at that location, calculated using a Gaussian kernel density estimate. }
\label{fig: SAR against GEDI}
\end{figure}

In northeastern region, cross-polarised SAR showed strong relationships with GEDI RH98 (Fig. \ref{fig: SAR against GEDI}a, c), while the relationships between co-polarised SAR and GEDI RH98 were not as expected (Fig. \ref{fig: SAR against GEDI}b, d). In contrast, in southwestern region, neither polarization showed a significant relationship with GEDI RH metrics, which might be substantially influenced by the steep slopes. There remains a substantial amount of GEDI data that does not meet the required quality standards. Generally, the backscatter of SAR from shorter trees is expected to be lower than that from taller trees. We therefore filtered the GEDI footprints in both regions using the relationships developed in northeastern regions, and only retained the GEDI data that fell within a range of plus and minus 3 dB from the fitted curve.

\subsubsection{AGB of GEDI footprints}
The GEDI L4B provides highly accurate biomass maps; however, there remains considerable uncertainty in biomass estimation for China \citep{dubayah2022gedi}. Furthermore, the reliability of the models used in GEDI L4A to convert RH metrics to AGB for Asian is relatively low, as evidenced by small R$^2$ values and large \%RMSE \citep{duncanson2022aboveground}. The GEDI L4 biomass products heavily rely on field data, and little data from China is included in the calibration process. Consequently, there is a pressing need to establish local models for calculating AGB from RH metrics, using field data collected within China.

\begin{figure}[H]
    \centering
        \centering
        \begin{tikzpicture}
            \node[anchor=south west,inner sep=0] (image) at (0,0) {\includegraphics[width=\textwidth]{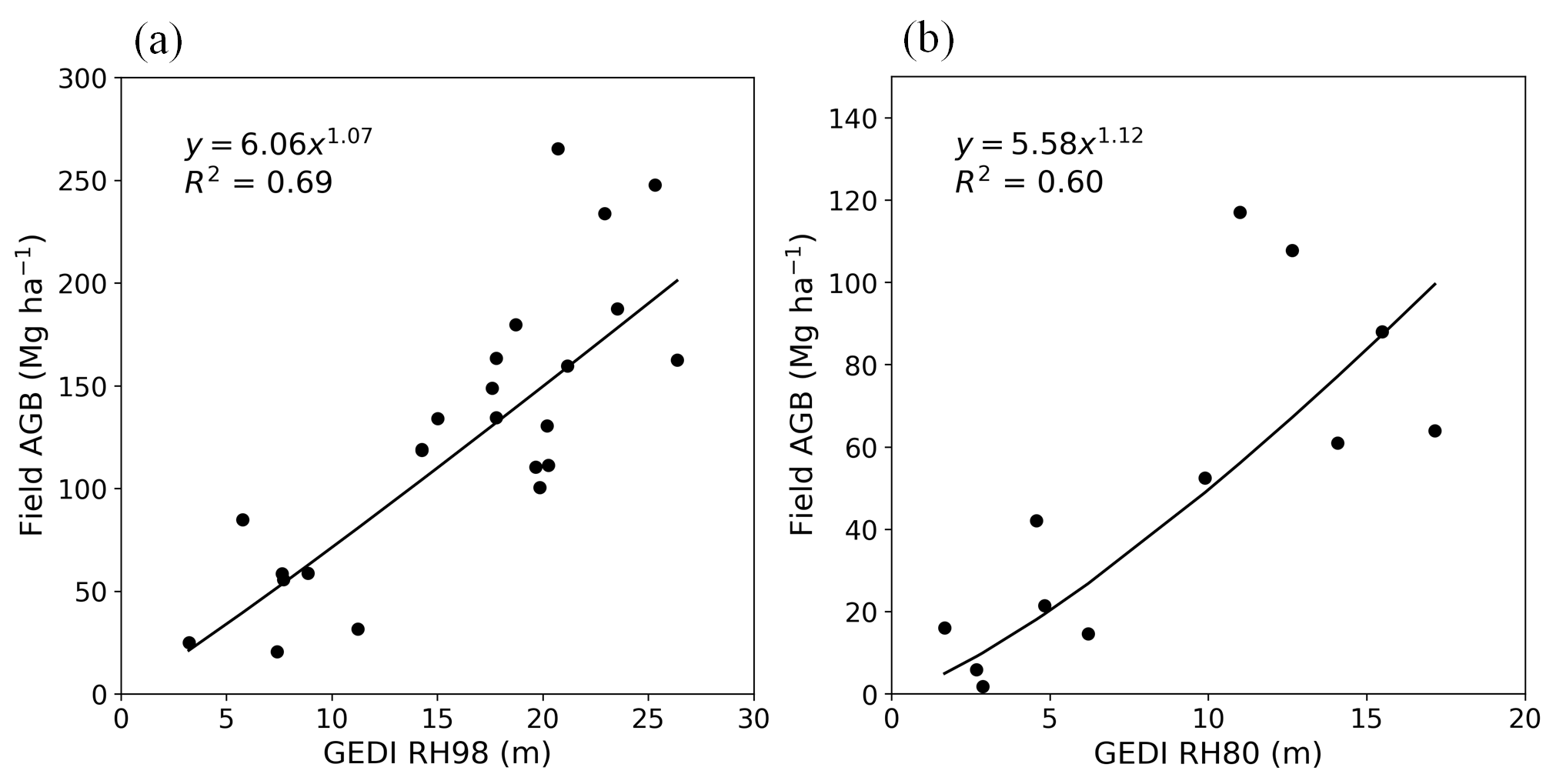}};
        \end{tikzpicture}
\caption{Field measured AGB are plotted against (a) RH98 in the northeastern region, and (b) RH80 in the southwestern region.}
\label{fig: Field AGB vs RH}
\end{figure}

In order to obtain AGB values for the GEDI footprints, we analysed the correlation between field AGB and RH values ranging from RH30 to RH98. In the northeastern region, the closest relationship was observed between field AGB and RH98, while in the mountainous southwestern region, field AGB exhibited a stronger correlation with RH80 (Fig. \ref{fig: Field AGB vs RH}). We converted RH98 and RH80 to AGB using the fitted relationships for both regions, respectively.

\subsubsection{Accuracy assessment}
We evaluated the model performance using a 5-fold cross-validation approach. The dataset was divided into five distinct subsets. Four subsets were employed for training, while the remaining one was used to test the model's accuracy. This procedure was repeated five times, ensuring each subset served as the test set once. Subsequent to these iterations, the results from all five test sets were aggregated to provide a comprehensive assessment of the model's accuracy. 

The model performance and the accuracy of the results were evaluated using R$^2$ and RMSE, computed using the following equations:
\begin{equation}
R^{2} = 1 - \frac{\sum_{i=1}^{n}(y_i - \hat{y}_i)^2}{\sum_{i=1}^{n}(y_i - \bar{y})^2}
\end{equation}
\begin{equation}
RMSE = \sqrt{\frac{1}{n} \sum_{i=1}^{n}(y_i - \hat{y}_i)^2}
\end{equation}
\begin{equation}
\mathrm{Bias} = \frac{1}{K} \sum_{i=1}^{K} (\hat{y}_i - y_i)
\end{equation}
where $y_i$ is the field measured AGB or the AGB derived from GEDI RH metrics, $\hat{y}_i$ is the model predicted value, $\bar{y}$ is the average value of $y_i$, and \( K \) is the cumulative number of observations.

\subsubsection{Wall-to-wall AGB map}
In this study, we extrapolated the extensive GEDI footprints into continuous AGB maps using LightGBM and random forest in Python 3.8. A comprehensive set of 25 remote sensing layers were employed in the generation of wall-to-wall maps. These layers comprised a wide array of data, including bands and ratios from SAR, optical bands, the maximum and average NDVI values, data from the DEM and slope, along with gridded latitude and longitude information.

\begin{figure}[H]
    \centering
    \begin{subfigure}{1.0\textwidth}
        \centering
        \begin{tikzpicture}
            \node[anchor=south west,inner sep=0] (image) at (0,0) {\includegraphics[width=\textwidth]{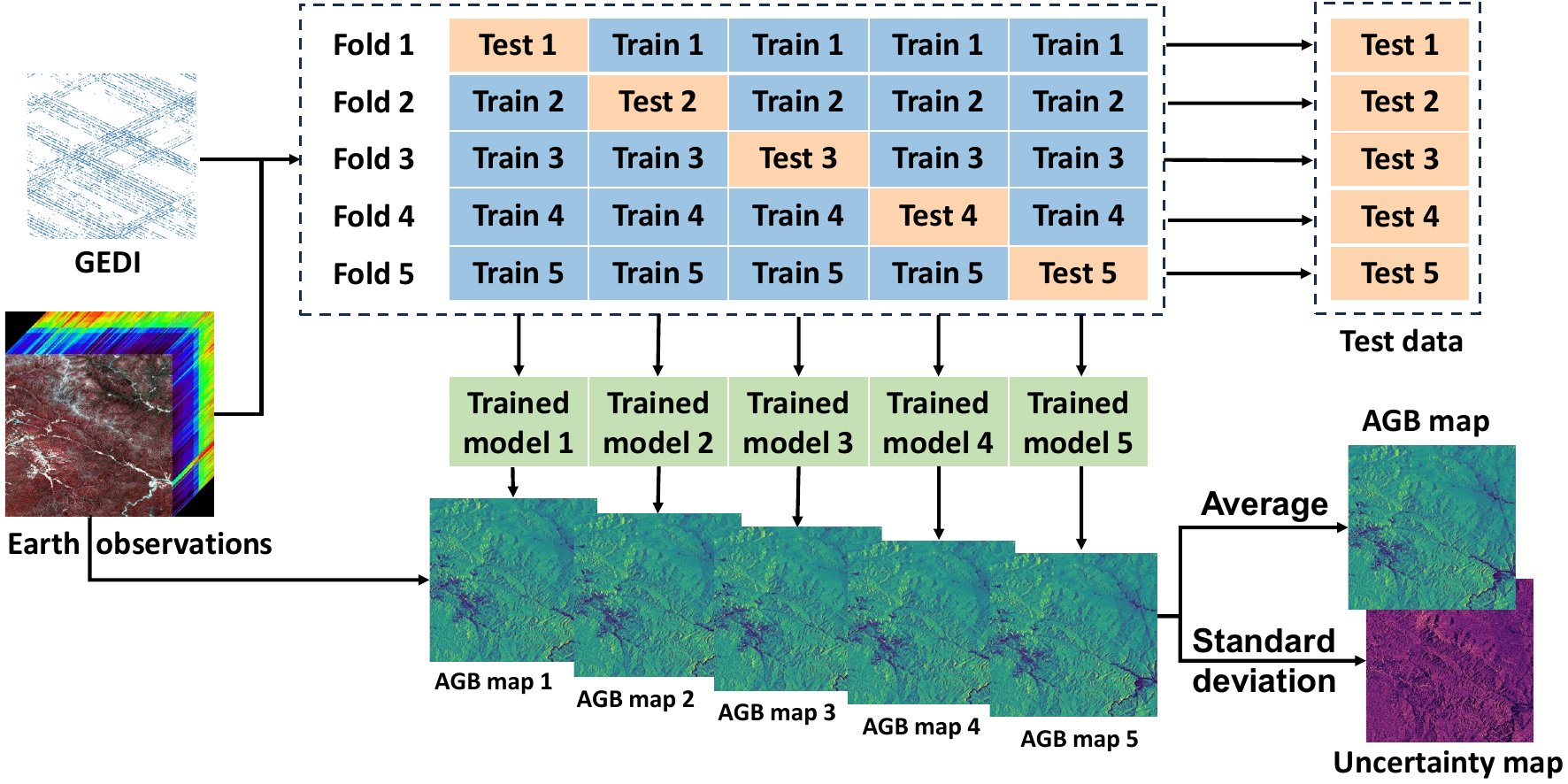}};
        \end{tikzpicture}
    \end{subfigure}%
\caption{Framework for estimating AGB and its uncertainty.}
\label{fig: Cross validation}
\end{figure}

Using the 5-fold cross-validation, each trained model from the five iterations was employed to predict the AGB maps across the study area, resulting in five separate AGB maps for each of the two regions. To derive a robust and consolidated AGB map, we computed the mean of these five predictions for each pixel. Additionally, to assess the variability and reliability of our predictions, we calculated the standard deviation across the five results for each pixel, which was then utilized to generate an uncertainty map (Fig. \ref{fig: Cross validation}).

To show the spatial distribution of forest AGB in both regions, the AGB maps were masked by a forest cover map in a 2021. The 2021 forest cover was derived from the Hansen et al forest map \citep{hansen2013high}, based on the 2000 forest cover map as a baseline. To generate the 2021 map, the degraded areas between 2000 and 2021 were excluded, and the forest areas that experienced an increase from 2000 to 2012 were added to the map (there is no information of forest gain after 2012). 

\section{Results}
\subsubsection{Model performance}
\begin{figure}[H]
    \centering
    \begin{subfigure}{1.0\textwidth}
        \centering
        \begin{tikzpicture}
            \node[anchor=south west,inner sep=0] (image) at (0,0) {\includegraphics[width=\textwidth]{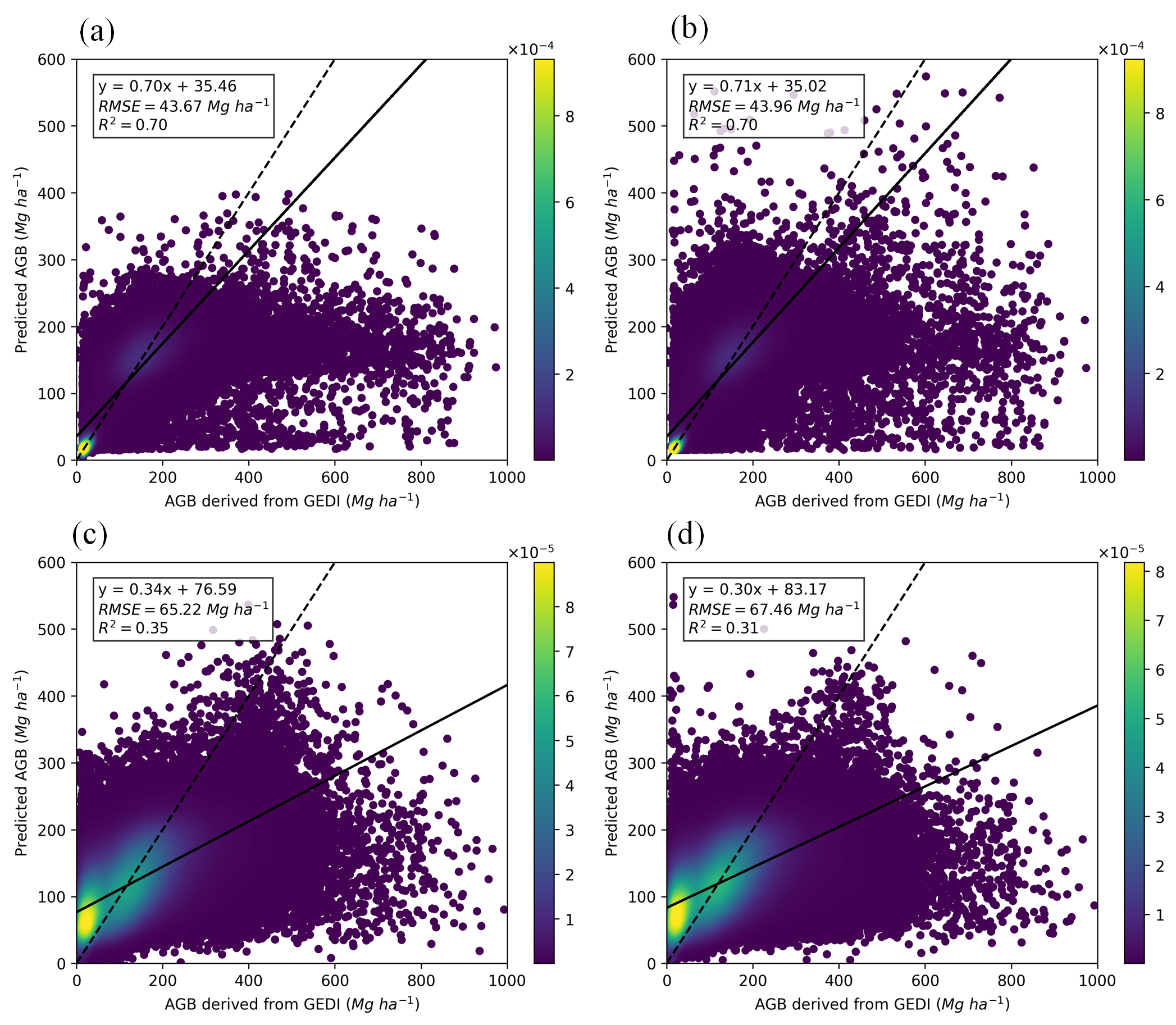}};
        \end{tikzpicture}
    \end{subfigure}%
\caption{The predicted AGB values against AGB derived from GEDI RH metrics before filtering using earth observation data. (a) LightGBM in northeastern region, (b)random forest in northeastern region, (c) LightGBM in southwestern region, (d) random forest in southwestern region. The color of each point in the scatter plot represents the estimated density of points at that location.}
\label{fig: model performance}
\end{figure}

Filtering GEDI data using parameters such as  quality\_flag, degrade, sensitivity, solar elevation (acquisition time), and beam intensity had been demonstrated to be an effective method for enhancing data quality. 1,405,451 and 2,042,330 GEDI footprints, filtered by the parameters above, were used to generate AGB maps for the northeastern and southwestern regions, respectively. The GEDI data was divided into  training and test datasets at a ratio of 6:4. Scatter plots of the estimates compared to the test data are shown in Fig. \ref{fig: model performance}, the variation of test data density is also illustrated. This initial analysis, which utilized GEDI footprints irrespective of GEDI RH metrics against SAR backscatter, indicated that both models exhibited strong performance in the northeastern region. However, neither model performed well in the southwestern region. There are many outliers in Fig. \ref{fig: model performance}, which may be caused by geolocation uncertainties in the GEDI data (Fig. 3 and section \ref{sec:GEDI data filtering}), or steep slopes.

\begin{figure}[H]
    \centering
    \begin{subfigure}{1.0\textwidth}
        \centering
        \begin{tikzpicture}
            \node[anchor=south west,inner sep=0] (image) at (0,0) {\includegraphics[width=\textwidth]{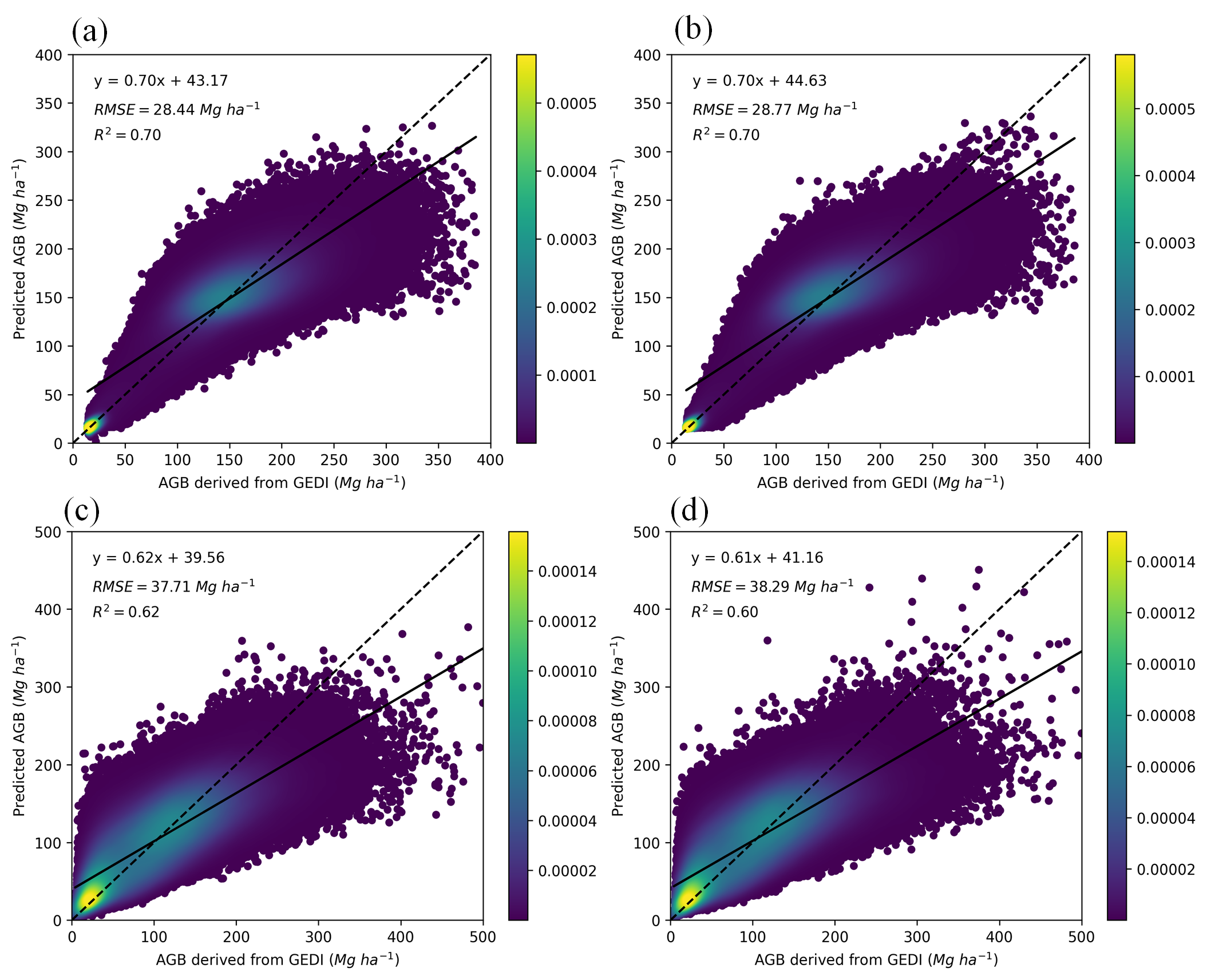}};
        \end{tikzpicture}
    \end{subfigure}%
\caption{The predicted AGB values against AGB derived from GEDI RH metrics after filtering using earth observation data. (a) LightGBM in northeastern region, (b)random forest in northeastern region, (c) LightGBM in southwestern region, (d) random forest in southwestern region. The color of each point in the scatter plot represents the estimated density of points at that location.}
\label{fig: model performance using filtered GEDI}
\end{figure}

We then utilized the GEDI footprints filtered using SAR data as described in section \ref{sec:GEDI data filtering}, with 689,013 and 673,744 footprints remaining for the northeastern and southwestern regions. We employed two different models to predict forest AGB. Predicted AGB from both models against test dataset of AGB derived from GEDI RH metrics are plotted in Fig. \ref{fig: model performance using filtered GEDI}. The performance of both models in the northeastern region did not show significant improvement, except for a slight reduction in the RMSE and a notable reduction in the number of outliers. However, both models showed significant improvement in the southwestern region. The R$^2$ and RMSE of the LightGBM  model improved from 0.35 and 65.22 Mg ha\textsuperscript{-1} to 0.62 and 37.71 Mg ha\textsuperscript{-1}, respectively. Similarly, the Random Forest model saw its R$^2$ and RMSE improve from 0.31 and 67.46 Mg ha\textsuperscript{-1} to 0.60 and 38.29 Mg ha\textsuperscript{-1}, respectively.

Both models demonstrated similar accuracy in the two regions. However, LightGBM outperforms Random Forest in terms of computational speed. Table \ref{tab:computation_time} presents the computational time required for training and prediction using the LightGBM and Random Forest models. The reduced computational time of LightGBM can be attributed to its optimized algorithm and efficient implementation. The notable difference in computational speed between LightGBM and Random Forest makes LightGBM a favorable choice for applications where efficient processing is crucial, especially when dealing with large-scale datasets.

\begin{table}[H]
  \centering
  \captionsetup{labelsep=period} 
  \caption{Computation time for LightGBM and random forest. NE represents the northeastern region, and SW represents the southwestern region.}
  \label{tab:computation_time}
  \begin{tabular}{cccc}
    \toprule
    \multirow{2}{*}{Study area} & \multirow{2}{*}{Model} & Number of & Computation \\
     &  & GEDI footprints & time (min) \\
    \midrule
    NE & LightGBM & 1,405,451 & 89 \\
    NE & random forest & 1,405,451 & 233 \\
    NE & LightGBM & 689,013 & 51 \\
    NE & random forest & 689,013 & 145 \\
    SW & LightGBM & 2,042,330 & 97 \\
    SW & random forest & 2,042,330 & 329 \\
    SW & LightGBM & 673,744 & 69 \\
    SW & random forest & 673,744 & 187 \\
    \bottomrule
  \end{tabular}
\end{table}

\subsubsection{Accuracy of the AGB maps}
\begin{figure}[H]
    \centering
    \begin{subfigure}{1.0\textwidth}
        \centering
        \begin{tikzpicture}
            \node[anchor=south west,inner sep=0] (image) at (0,0) {\includegraphics[width=\textwidth]{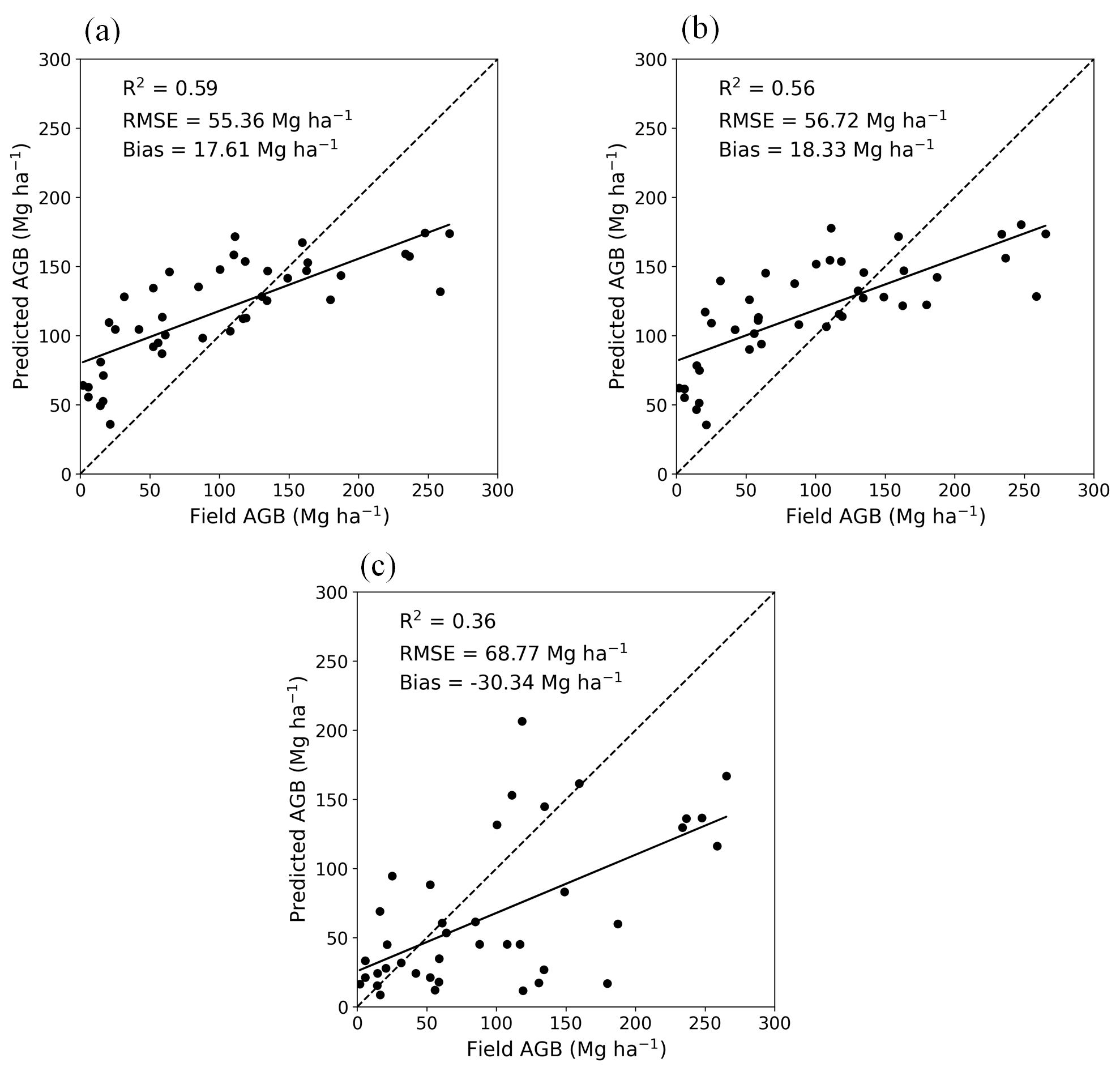}};
        \end{tikzpicture}
    \end{subfigure}%

\caption{Scatter plot between predicted AGB and field measured AGB. (a) LightGBM, (b)random forest, (c) GEDI L4B (Version 2). The solid line represent the fitted line, and the dotted line represents 1:1 line.}
\label{fig: Accuracy of AGB maps}
\end{figure}

The accuracy of the AGB maps produced by both models was evaluated using 42 field plots. The estimated AGB from LightGBM and random forest correspond well with the field AGB in both regions (Fig. \ref{fig: Accuracy of AGB maps}). Overall, the accuracy of LightGBM is slightly higher than that of Random Forest. The AGB map generated by LightGBM exhibits an R$^2$ of 0.59, an RMSE of 55.36 Mg ha\textsuperscript{-1}, and a bias of 17.61 Mg ha\textsuperscript{-1}. In contrast, the AGB map produced by the random forest model yielded an R$^2$ of 0.56, an RMSE of 56.72 Mg ha\textsuperscript{-1}, and a bias of 18.33 Mg ha\textsuperscript{-1}. Compared to the field data, both models have overestimated biomass in low biomass areas and underestimated biomass in high biomass areas, as is common in biomass mapping using machine learning \citep{shendryk2022fusing, huang2019integration}. This trend is also evident when evaluating model performance using AGB derived from the GEDI footprints (Fig. \ref{fig: model performance using filtered GEDI}).

For comparison, we also conducted an analysis using collected field data against the 1 km resolution GEDI L4B Version 2 AGB map. Due to gaps in the GEDI L4B data product, three field plots did not overlap with the GEDI L4B data. Consequently, we conducted the accuracy assessment using the remaining 39 field plots. The results indicated an R$^2$ of 0.36, an RMSE (Root Mean Square Error) of 68.77 Mg ha\textsuperscript{-1}, and a bias of -30.34 Mg ha\textsuperscript{-1}.

\subsubsection{Spatial distribution of AGB in both regions}
\begin{figure}[H]
    \centering
    \begin{subfigure}{1.0\textwidth}
        \centering
        \begin{tikzpicture}
            \node[anchor=south west,inner sep=0] (image) at (0,0) {\includegraphics[width=\textwidth]{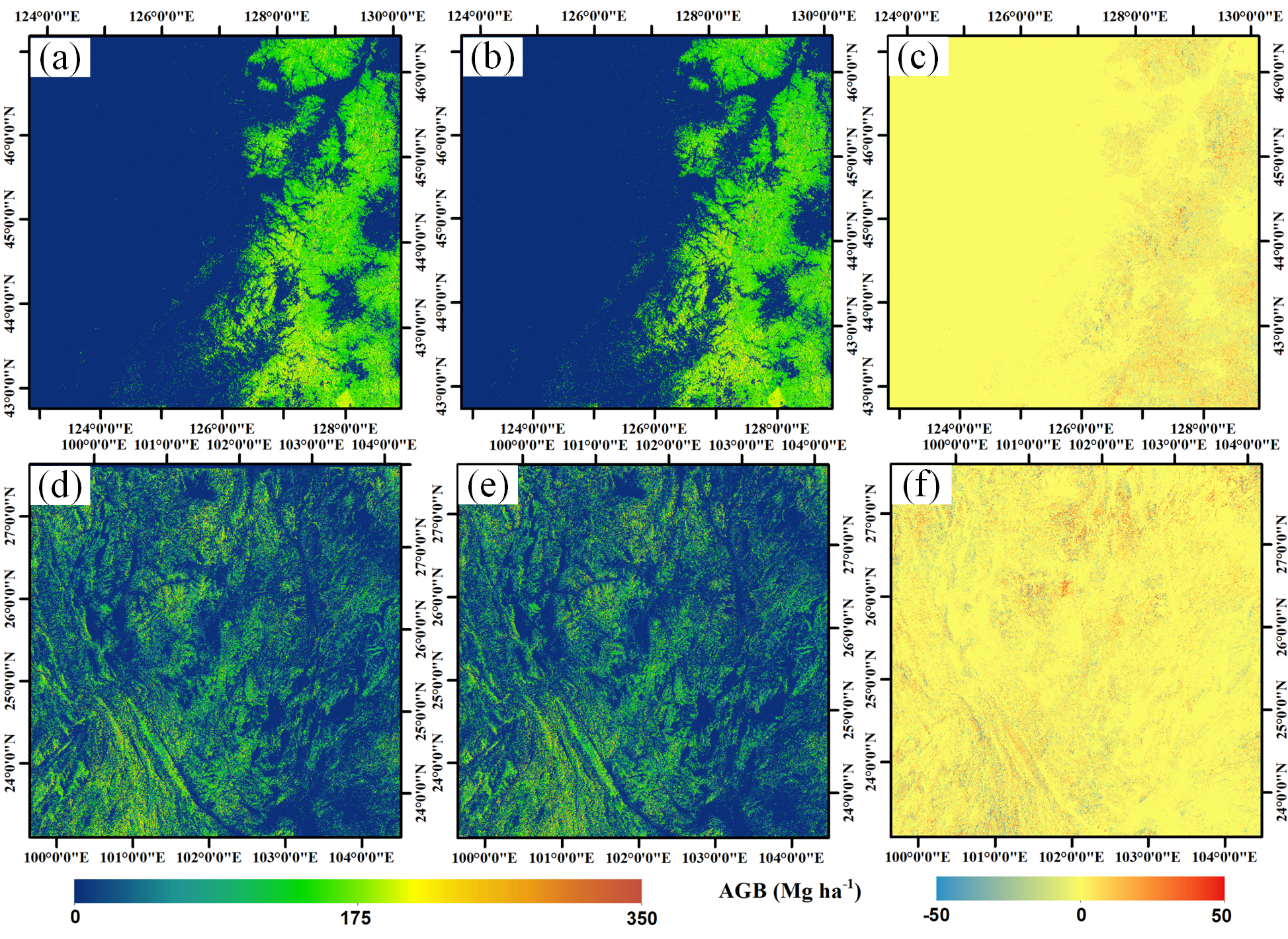}};
        \end{tikzpicture}
    \end{subfigure}%
\caption{Spatial distribution of AGB and the differences between the AGB maps from the two models at 25 m resolution. (a) LightGBM in the northeastern region, (b) Random forest in the northeastern region, (c) Difference Map between LightGBM and RF Estimated AGB in the northeastern region (d) LightGBM in the southwestern region, (e) Random forest in the southwestern region, (f) Difference Map between LightGBM and RF Estimated AGB in the southwestern region.}
\label{fig: AGB_distribution_TwoRegions}
\end{figure}

\begin{figure}[H]
    \centering
    \begin{subfigure}{1\textwidth}
        \centering
        \begin{tikzpicture}
            \node[anchor=south west,inner sep=0] (image) at (0,0) {\includegraphics[width=\textwidth]{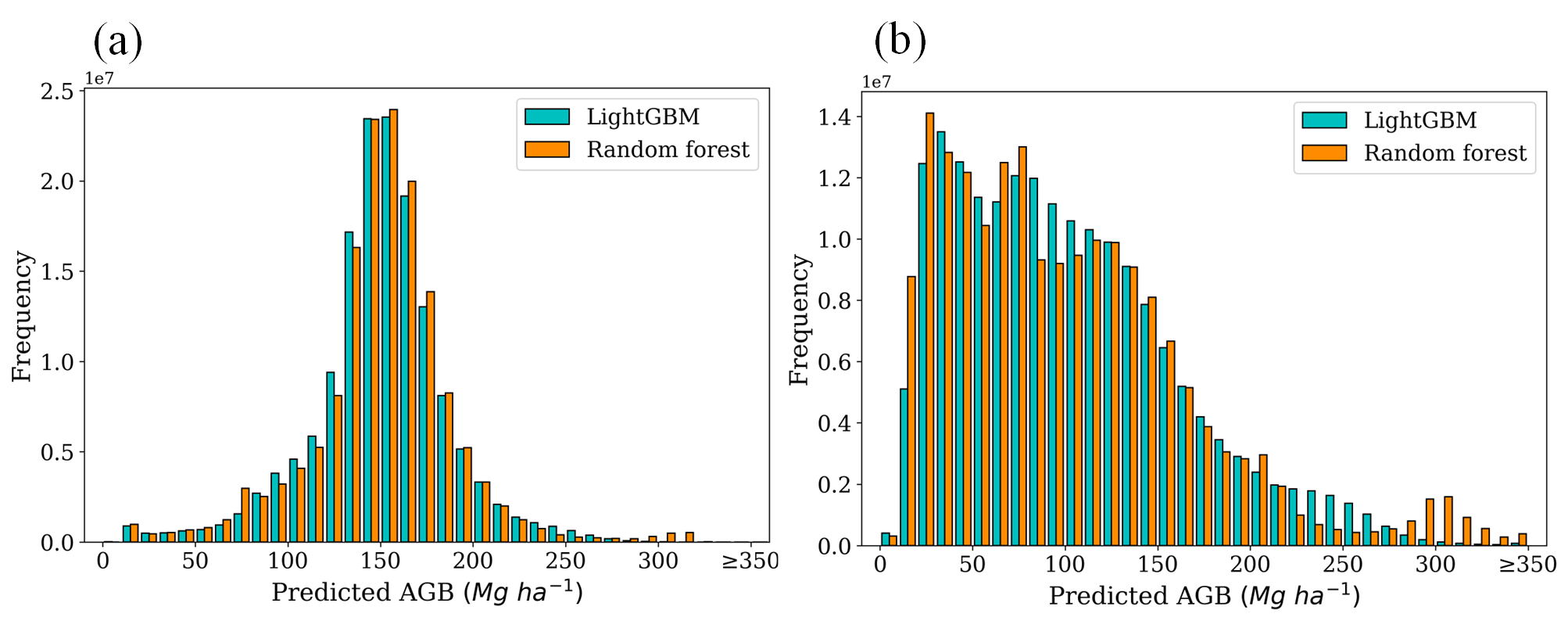}};
            \node[rectangle, draw=none, fill=none, inner sep=1pt, anchor=north west, font=\sffamily, xshift=17pt, yshift=14pt] at (image.north west) {(a)};
        \end{tikzpicture}
    \end{subfigure}
\caption{Frequency distribution histograms of AGB, in 10 Mg ha\textsuperscript{-1}, for (a)The northeastern region, (b)The southwestern region.}
\label{fig: Histograms}
\end{figure}

\begin{figure}[H]
    \centering
    \begin{subfigure}{1.0\textwidth}
        \centering
        \begin{tikzpicture}
            \node[anchor=south west,inner sep=0] (image) at (0,0) {\includegraphics[width=\textwidth]{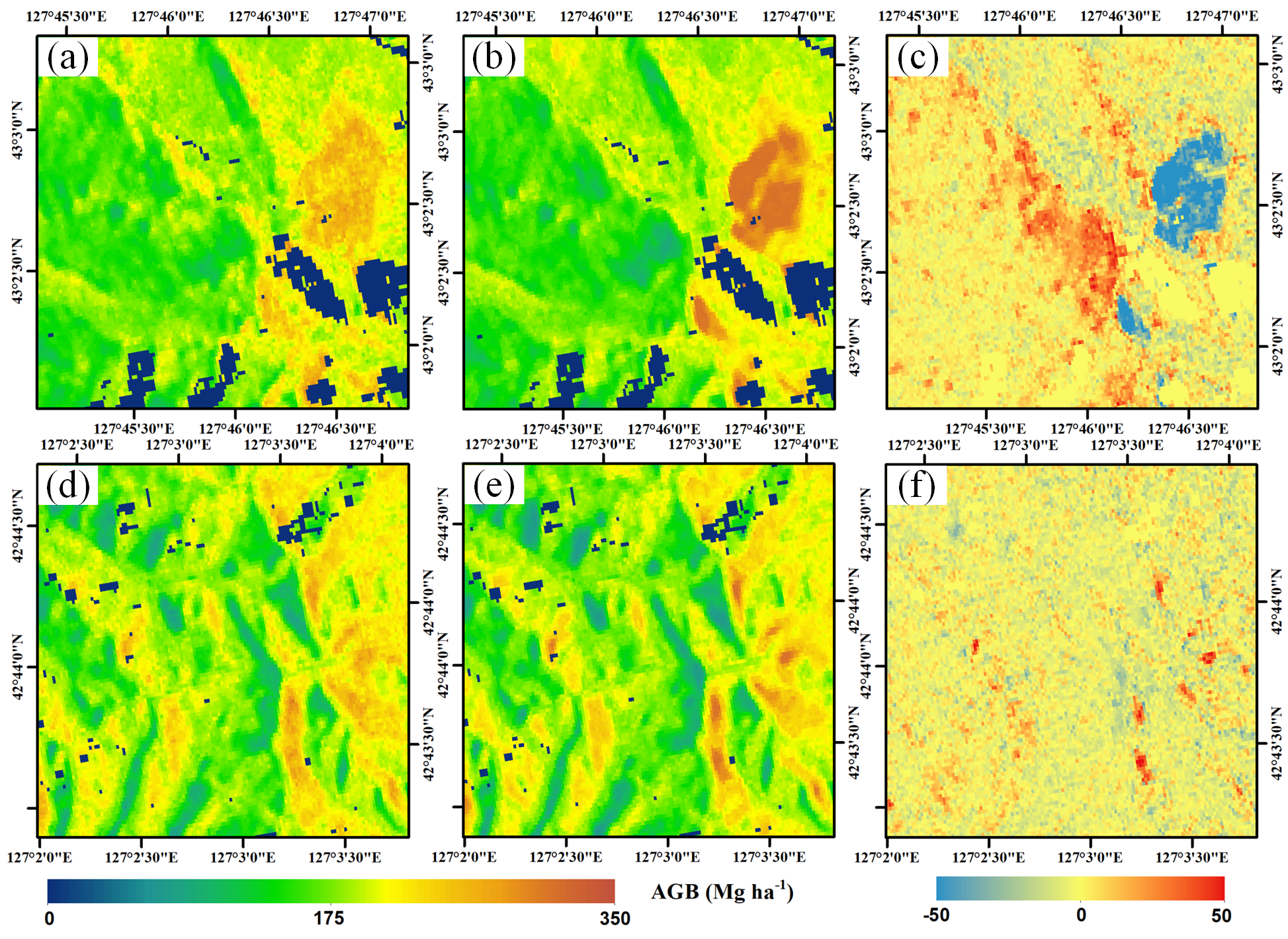}};
        \end{tikzpicture}
    \end{subfigure}%
\caption{Spatial distribution of AGB for two 2500 m x 2500 m areas. (a) (d) LightGBM, (b) (e) Random forest, (c) (f) difference maps.}
\label{fig: AGB_distribution_SmallArea}
\end{figure}

The wall-to-wall forest AGB maps by LightGBM and random forest in both regions are shown in Fig. \ref{fig: AGB_distribution_TwoRegions}. In the northeastern region, high AGB values are predominantly concentrated in the eastern part of the study area. Conversely, in the southwestern region, the distribution of AGB exhibits a more dispersed pattern. In both the northeast and southwest regions, the spatial distribution patterns of AGB predicted by the LightGBM and Random Forest models show strong agreement, perhaps unsurprisingly given they are based on the same Earth Observation data inputs. This is further evidenced in the histograms, where both models generate similar overall AGB distributions \ref{fig: Histograms}. In the northeastern region, the mean difference in AGB estimates between LightGBM and Random Forest was -1.59 Mg ha\textsuperscript{-1}. Similarly, in the southwestern region, the mean AGB estimate difference was -1.44 Mg ha\textsuperscript{-1}. Although the overall spatial distribution patterns show strong agreement, difference maps in Fig. \ref{fig: AGB_distribution_TwoRegions} reveal there are still discernible differences between the LightGBM and Random Forest AGB predictions. To better illustrate these spatial differences at a finer scale, zoomed-in 2500 m x 2500 m maps are presented in Fig. \ref{fig: AGB_distribution_SmallArea}. Focusing on this local scale enables better visualization and analysis of the details and spatial heterogeneity in the AGB distribution. The AGB maps generated by the Random Forest model exhibit a deeper characterization of extreme values (Fig. \ref{fig: AGB_distribution_SmallArea}b, e), both at the high and low ends of the AGB range. In contrast, the AGB maps generated by the LightGBM model display a relatively smoother transition between different AGB values (Fig. \ref{fig: AGB_distribution_SmallArea}a, d), with more gradual change.

\begin{figure}[H]
    \centering
    \begin{subfigure}{1.0\textwidth}
        \centering
        \begin{tikzpicture}
            \node[anchor=south west,inner sep=0] (image) at (0,0) {\includegraphics[width=\textwidth]{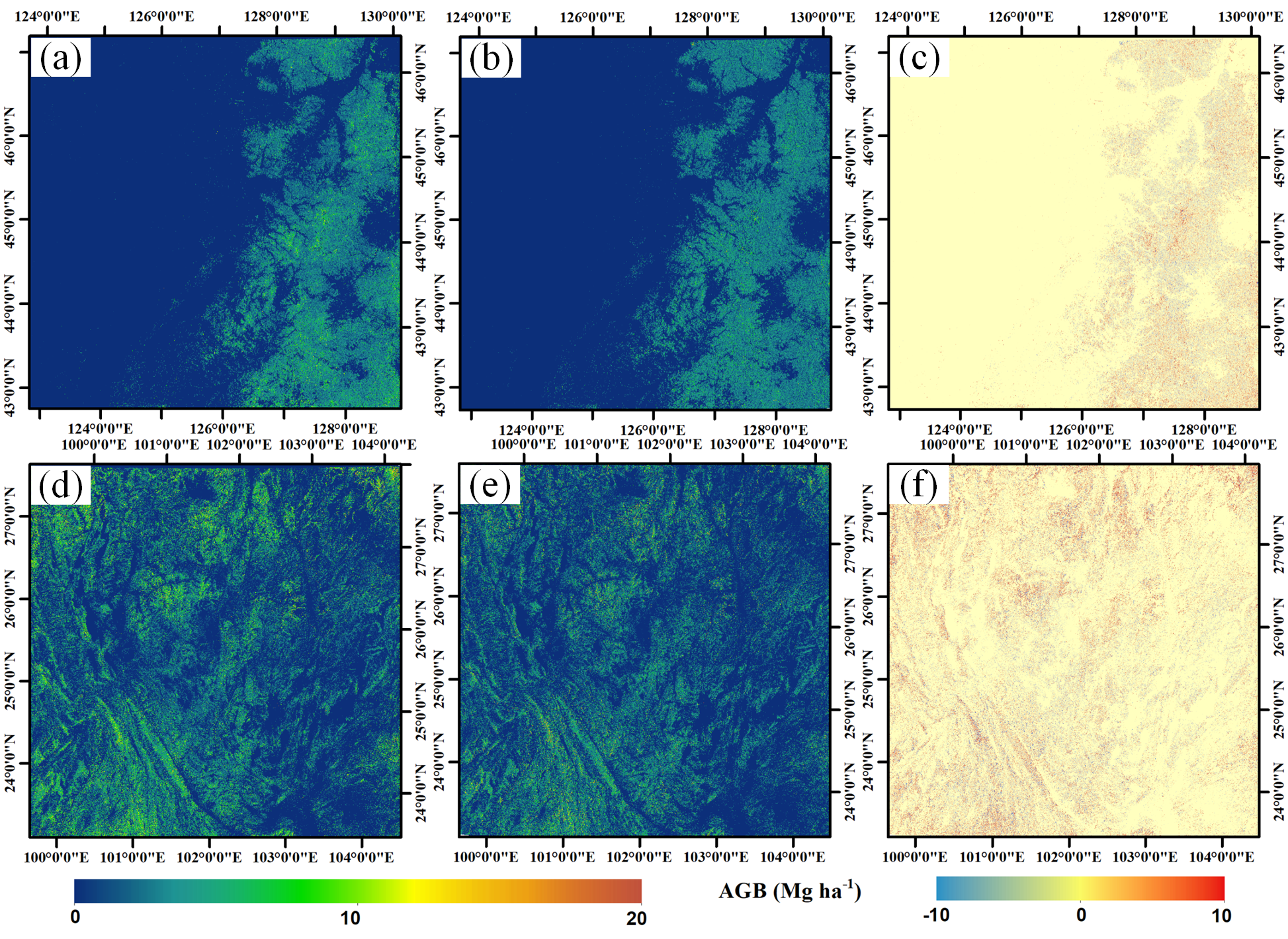}};
        \end{tikzpicture}
    \end{subfigure}%
\caption{Spatial distribution of uncertainty and the differences between the uncertainty maps from the two models at 25 m resolution. (a) LightGBM in the northeastern region, (b) Random forest in the northeastern region, (c) Difference Map between LightGBM and RF uncertainty in the northeastern region, (d) LightGBM in the southwestern region, (e) Random forest in the southwestern region, (f) Difference Map between LightGBM and RF uncertainty in the southwestern region.}
\label{fig: Uncertainty_distribution_TwoRegions}
\end{figure}

We also generated uncertainty maps based on the standard deviation of results from 5-fold cross-validation (Fig. \ref{fig: Uncertainty_distribution_TwoRegions}). Both LightGBM and Random Forest showed relatively low uncertainty in the two study regions. For LightGBM, the uncertainty in the northeastern region was 3.28 Mg ha\textsuperscript{-1}, while in the southwestern region it was 4.22 Mg ha\textsuperscript{-1}. Meanwhile, Random Forest uncertainty was 2.96 Mg ha\textsuperscript{-1} in the northeast and 3.44 Mg ha\textsuperscript{-1} in the southwest. However, LightGBM exhibited slightly higher uncertainty, indicating it may be more sensitive to differences across folds in the training data. Further analysis is warranted to determine the factors contributing to the uncertainty patterns.

\section{Discussion}
\subsection{Influence of steep slopes}
\begin{figure}[H]
    \centering
    \begin{subfigure}{1.0\textwidth}
        \centering
        \begin{tikzpicture}
            \node[anchor=south west,inner sep=0] (image) at (0,0) {\includegraphics[width=\textwidth]{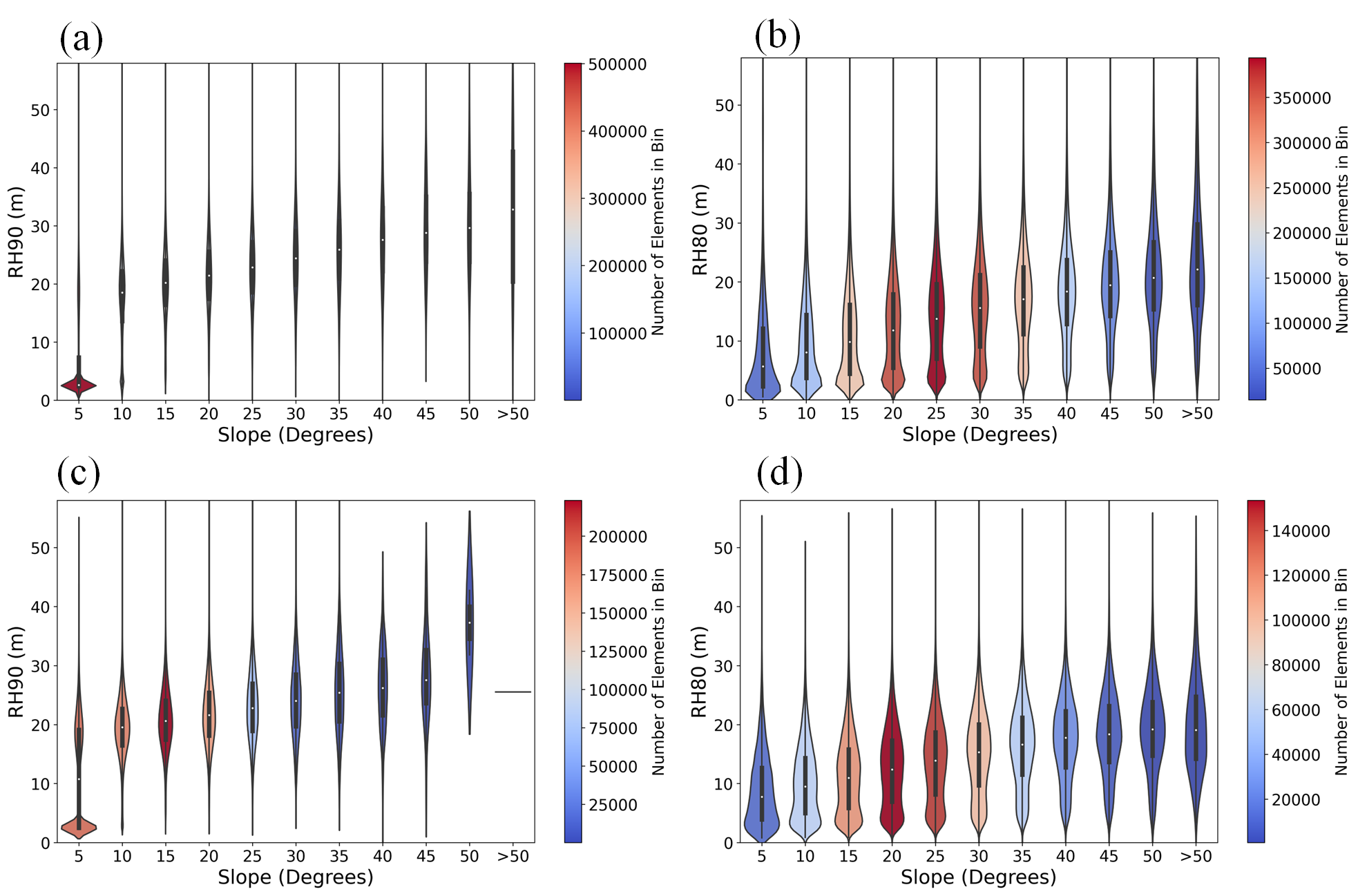}};
        \end{tikzpicture}
    \end{subfigure}%
\caption{GEDI RH metrics are plotted against slope. The slope bin of 5 indicates slope from 0 to 5, and similar with all other bins. The colors of the boxes represent the quantity of GEDI footprints in different slope bins. (a) RH98 in northeastern region, (b) RH80 in southwestern region, (c) RH98 in northeastern region after filtering using SAR data, (d) RH80 in southwestern region after filtering using SAR data.}
\label{fig: RH VS Slope}
\end{figure}

Both regions are characterized by a significant amount of mountainous terrain, resulting in many GEDI footprints falling in areas with substantial slopes, especially in the southwestern region (Fig. \ref{fig: RH VS Slope}). This could potentially impact the results, as steep slopes could lead to overestimation or underestimation of canopy height by GEDI data, depending on the steepness of the slope and spatial distribution of trees within footprints \citep{chen2010retrieving, schneider2020towards}. Slopes also impact the backscatter of the SAR data \citep{shimada2011generation}. Ideally the SAR data should be corrected for increased or decreased backscatter caused by terrain slopes. However, the relationship between slope and backscatter is complex and depends on forest structure and properties, so slope corrections are not always feasible \citep{ramachandran2021evaluation}. Additionally, some steep slopes can cause SAR layover and shadow effects, resulting in no useful SAR data for those areas \citep{samuele2021mapping}. These terrain factors introduce uncertainties in both the GEDI lidar and SAR data which should be considered, especially in regions with substantial topography. 

As the slope of the GEDI footprints increases, there is a noticeable decrease in the R$^2$ values, and the RMSE gradually increases (Fig. \ref{fig: R2, RMSE VS Slope}) in both regions. Overall, the two models demonstrated similar accuracy in flat areas. However, in areas with steep slopes, LightGBM outperformed the random forest model, with a larger R$^2$ and smaller RMSE. We chose not to remove GEDI footprints from areas with steep slopes, because their removal could lead to  unreliable AGB predictions in these steep areas, as much remaining forest is genuinely located on steep slopes in these regions. In the northeastern region, the average slope of GEDI is 10.23 degrees, and for the GEDI footprints within the Hansen forest mask, it is slightly higher at 11.97 degrees. The average slope of GEDI in the northeastern region measures 20.38 degrees, whereas the average slope of GEDI footprints within the Hansen forest mask stands at 21.18 degrees. More field data collected in areas with steep slopes are needed to evaluate the impact of removing GEDI footprints in steep areas on the accuracy of AGB estimation in these areas.

\begin{figure}[H]
    \centering
    \begin{subfigure}{1.0\textwidth}
        \centering
        \begin{tikzpicture}
            \node[anchor=south west,inner sep=0] (image) at (0,0) {\includegraphics[width=\textwidth]{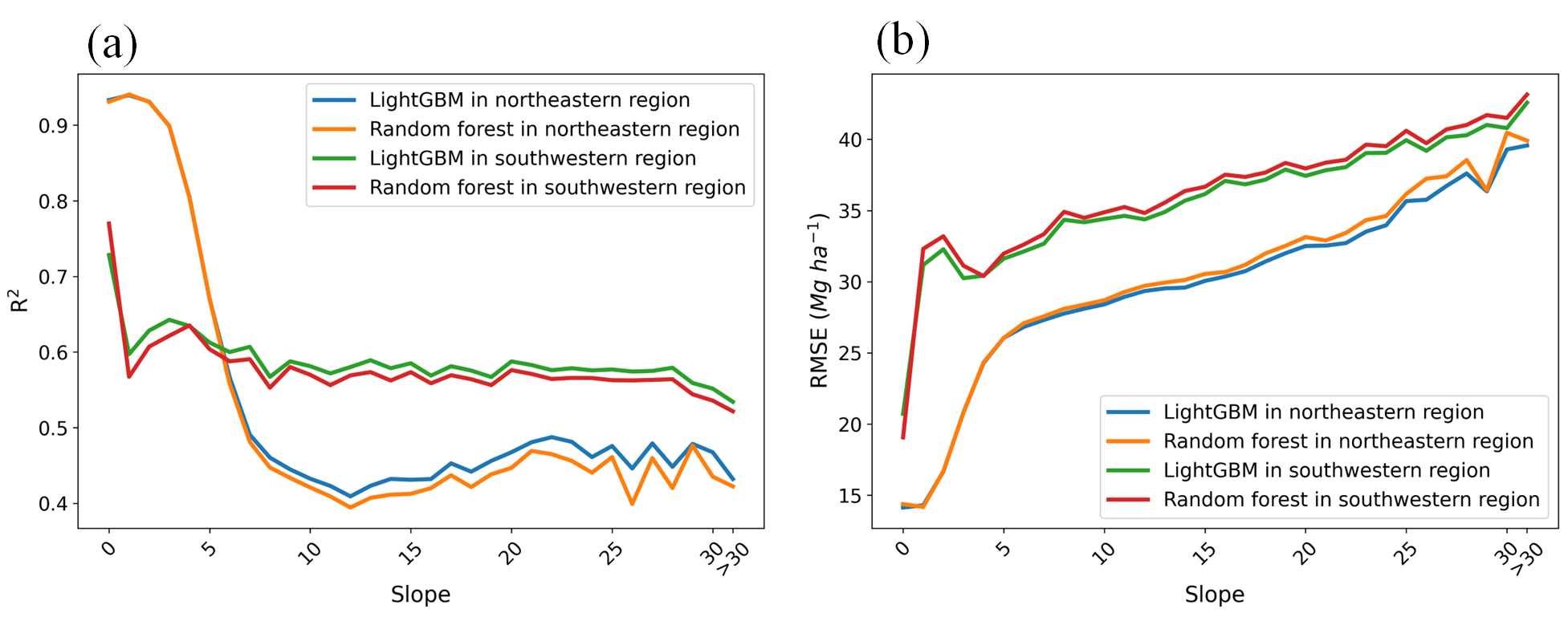}};
        \end{tikzpicture}
    \end{subfigure}%
\caption{Model performance in both regions are plotted against slope.}
\label{fig: R2, RMSE VS Slope}
\end{figure}

\subsection{Scalability of the approach}
\begin{figure}[H]
    \centering
    \begin{subfigure}{1.0\textwidth}
        \centering
        \begin{tikzpicture}
            \node[anchor=south west,inner sep=0] (image) at (0,0) {\includegraphics[width=\textwidth]{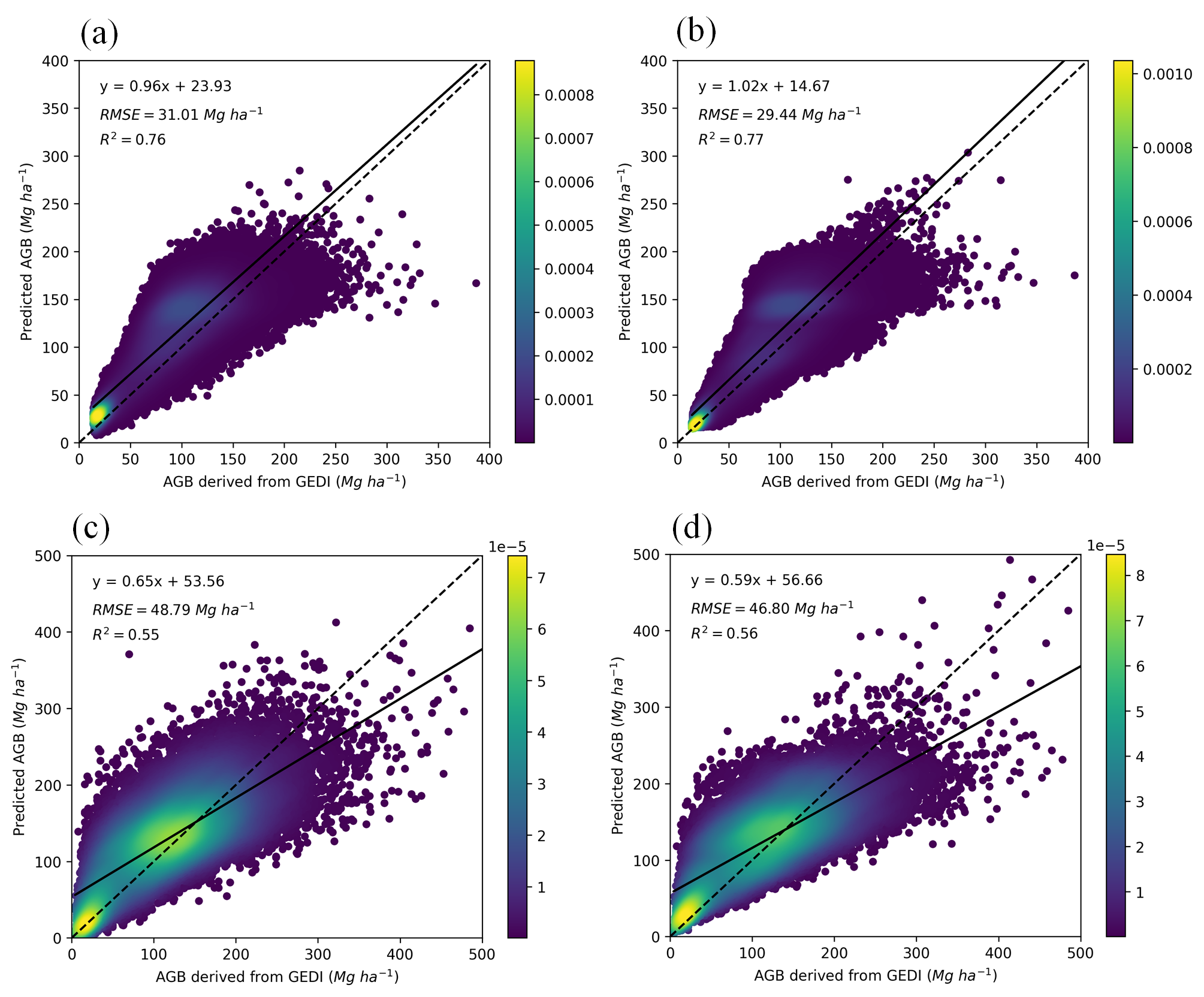}};
        \end{tikzpicture}
    \end{subfigure}%
\caption{Performance of trained models in other area. (a) Trained LightGBM in the northern test region, (b) Trained random forest in the northern test region, (c) Trained LightGBM in the middle southern region, (d) Trained random forest in the middle southern region. The color of each point in the scatter plot represents the estimated density of points at that location.}
\label{fig: model performance in other area.}
\end{figure}

The models trained with GEDI and remote sensing imagery in the northeastern and southwestern regions were applied directly to two 100 x 100 km areas (Fig. \ref{fig: Study area}). These test areas are located 90 km from the northeastern region and 450 km from the southwestern region, respectively. In the northern test area, the results of both models closely aligned with the AGB derived from GEDI data (Fig. \ref{fig: model performance in other area.}). The LightGBM and random forest models achieved R$^2$ values of 0.76 and 0.77, with respective RMSE values of 31.01 and 29.44 Mg ha\textsuperscript{-1}. The fitted line and the 1:1 line between the model predictions and the AGB derived from GEDI were very close, indicating no significant saturation point within the AGB range below 300 Mg ha\textsuperscript{-1}. In the middle southern test area, the R$^2$ values were slightly lower, with LightGBM and random forest models achieving R$^2$ values of 0.55 and 0.56, respectively. However, the RMSE values remained relatively low, at 48.79 and 46.80 Mg ha\textsuperscript{-1}, respectively, indicating high accuracy in this area as well.

\subsection{Linking field AGB and GEDI RH metrics}
In this study, we measured canopy height and DBH of trees in 42 field plots, with 36 of them co-located with GEDI footprints. There are various allometric equations to estimate forest AGB using height and DBH, depending on tree species and regions. The AGB estimates derived from different allometric equations can vary significantly. In this study, we used a general allometric equation developed for the trees in China \citep{luo2018china}. The selection of field plots significantly influence the conversion of GEDI RH metrics into AGB. It is crucial that the selection of field plots is representative. To this end, we have endeavored to ensure that our field points are evenly distributed spatially, thereby enhancing their representativeness.

\subsection{Geolocation uncertainty of GEDI footprint}
In this study, we used GEDI L2A version 2 dataset, which improved geolocation
accuracy compared to version 1. However, upon comparison with ALS data, we found that the geolocation uncertainty in the GEDI data remains substantial. This may introduce errors into the AGB estimation of this study from two perspectives. Firstly, our field plots and GEDI footprints are at the same location and of the same size. However, there may be a discrepancy of about ten meters with the actual location of the GEDI footprints. This discrepancy could potentially result in the relationship we've established between field AGB and GEDI RH not being as accurate as intended. The effect had been somewhat mitigated by our strategy for selecting field points, which ensures that our field points were at least 50 meters away from non-forest areas. In addition, the geolocation uncertainty  may introduce errors when matching GEDI footprints with continuous remote sensing images in the models. The geolocation uncertainty of GEDI footprints can be greatly improved through the use of ALS data. However, ALS data is scarce and expensive to acquire. Our research provides a reference for regions and countries where ALS data is lacking, especially for large regions. Our method of using known relationships between AGB and key remote sensing metrics to filter out GEDI footprints that are far from the expected values is shown here to be a reliable way to remove low reliability GEDI footprints, whatever the cause of error.

\section{Conclusions}
In this study, we presented an experiment designed to determine the universality of remote sensing-based forest biomass mapping approaches at 25 m resolution, using new field plots co-located with GEDI footprints in two contrasting regions in northeastern and southwestern China. Both LightGBM and random forest performed well in the two regions compared to the GEDI validation data and field AGB. We also applied the models trained in these two contrasting regions to a northern area and a central-southern area, respectively, and achieved good results.Our preliminary test of the scalability of the models across different areas seems to demonstrate their robustness and ability to capture the relationships between the GEDI data and continuous remote sensing data in different areas with similar forest types. The promising outcomes can be attributed to several key factors. Firstly, the availability of a large volume of GEDI footprints and multiple sources of remote sensing imagery, including SAR and optical data, contributes to the accuracy and comprehensiveness of the prediction. In addition, the stringent removal of poor-quality data from the GEDI dataset enhances the reliability of the results. Our study demonstrates the significant potential of using GEDI data to produce high resolution AGB maps of China in the GEDI era (2019 and onwards). 

LightGBM and random forest models exhibit similar performance in terms of accuracy. However, LightGBM outperforms random forest in terms of computational speed. With the growing number of space-borne sensors acquiring imagery and LiDAR footprints at high resolutions and frequencies, LightGBM's computational speed advantage empowers researchers to efficiently process large-scale and high-frequency AGB estimation (i.e. national scale annual estimates).

\section{Acknowledgements}
Edward Mitchard’s time was part funded by NERC grant SEOSAW (Grant number NE/P008755/1). The field work was supported by Davis Expedition Fund, Elizabeth Sinclair Irvine Bequest and Centenary Agroforestry 89 Fund, Moray Endowment Fund and Meiklejohn fund. The authors would like to thank ESA and JAXA for providing the Sentinel-1, Sentinel-2 and ALOS-2 PALSAR-2 mosaic data. 

\bibliographystyle{elsarticle-harv} 
\bibliography{references}





\end{document}